\documentclass[nohyperref]{article}
\usepackage[accepted]{icml2024}

\newcount\Comments 
\usepackage{microtype}
\usepackage{graphicx}
\usepackage{subfigure}
\usepackage{booktabs} 
\usepackage{etoolbox}
\usepackage{resizegather}
\usepackage{fancyvrb}

\usepackage{hyperref}
\usepackage[makeroom]{cancel}

\usepackage{amsmath}
\usepackage{amssymb}
\usepackage{mathtools}
\usepackage{amsthm}
\usepackage{bm}
\usepackage{eqparbox} 
\usepackage{setspace}

\DeclareMathSymbol{\shortminus}{\mathbin}{AMSa}{"39}

\let\classAND\AND
\let\AND\relax
\usepackage{algorithmic}

\let\AND\classAND
\AtBeginEnvironment{algorithmic}{\let\AND\algoAND}
\usepackage[capitalize,noabbrev]{cleveref}
\usepackage{multirow}
\usepackage{enumitem}
\usepackage{caption}
\usepackage{color}
\usepackage{colortbl}
\usepackage{xcolor}
\definecolor{crimson}{HTML}{FF0000}

\definecolor{tablecolor}{rgb}{0.8,0.8,0.8}

\newcommand{\barloss}{\bar{\ell}}

\newcommand{\loss}{\ell}

\newcommand{\vparam}{\vtheta}

\newcommand{\minibatch}{\mathcal{B}}

\newcommand\cut[1]{}

\newcommand{\squishlist}{
   \begin{list}{$\bullet$}
    { \setlength{\itemsep}{0pt}      \setlength{\parsep}{3pt}
      \setlength{\topsep}{3pt}       \setlength{\partopsep}{0pt}
      \setlength{\leftmargin}{1.5em} \setlength{\labelwidth}{1em}
      \setlength{\labelsep}{0.5em} } }

\newcommand{\squishlisttwo}{
   \begin{list}{$\bullet$}
    { \setlength{\itemsep}{0pt}    \setlength{\parsep}{0pt}
      \setlength{\topsep}{0pt}     \setlength{\partopsep}{0pt}
      \setlength{\leftmargin}{2em} \setlength{\labelwidth}{1.5em}
      \setlength{\labelsep}{0.5em} } }

\newcommand{\squishend}{
    \end{list}  }

{}
{}
{}

\newcommand{\half}{\mbox{$\frac{1}{2}$}}
\newcommand{\real}{\mbox{$\mathbb{R}$}}

\newcommand{\gauss}{\mbox{${\cal N}$}}

\newcommand{\softmax}{\calS}

\newcommand{\myvec}[1]{\mbox{$\mathbf{#1}$}}
\newcommand{\myvecsym}[1]{\mbox{$\boldsymbol{#1}$}}

\newcommand{\vepsilon}{\mbox{$\myvecsym{\epsilon}$}}

\newcommand{\vtheta}{\mbox{$\myvecsym{\theta}$}}

\newcommand{\vsigma}{\mbox{$\myvecsym{\sigma}$}}

\newcommand{\ve}{\mbox{$\myvec{e}$}}

\newcommand{\vf}{\mbox{$\myvec{f}$}}
\newcommand{\vg}{\mbox{$\myvec{g}$}}
\newcommand{\vh}{\mbox{$\myvec{h}$}}

\newcommand{\vm}{\mbox{$\myvec{m}$}}

\newcommand{\vv}{\mbox{$\myvec{v}$}}

\newcommand{\vy}{\mbox{$\myvec{y}$}}

\newcommand{\vI}{\mbox{$\myvec{I}$}}

\newcommand{\vV}{\mbox{$\myvec{V}$}}

\newcommand{\calS}{\mbox{${\cal S}$}}

\newcommand{\revision}[1]{{#1}}

\newcommand{\vddl}{\mathbf{g}_h}
\renewcommand{\vparam}{\bm{\theta}}
\renewcommand{\vI}{\mathbf{I}}

\crefname{section}{Sec.}{Sec.}
\crefname{thm}{Thm.}{Theorem}
\crefname{appendix}{App.}{Appendices}
\crefname{algorithm}{Alg.}{Algorithms}
\crefname{equation}{Eq.}{Eqs.}
\crefname{figure}{Fig.}{Figs.}
\creflabelformat{equation}{#2\textup{#1}#3} 

\theoremstyle{plain}

\theoremstyle{definition}

\theoremstyle{remark}

\usepackage[textsize=tiny]{todonotes}

\newcommand{\bbE}{\mathbb{E}}

\newcommand{\cL}{\mathcal{L}}

\DeclareMathOperator{\diag}{diag}

\DeclarePairedDelimiterX{\infdivx}[2]{{}}{{}}{%
	\left( #1\,\delimsize\|\,#2\right)%
}

\DeclareMathOperator{\KLop}{KL}
\newcommand{\myKL}{\mathbb{D}_{\KLop}\infdivx}

\title{Variational Learning is Effective for Large Deep Networks}

\begin{document}
\twocolumn[
\icmltitle{Variational Learning is Effective for Large Deep Networks}



\icmlsetsymbol{equal}{*}

\begin{icmlauthorlist}
\icmlauthor{Yuesong Shen}{equal,tum}
\icmlauthor{Nico Daheim}{equal,tud}
\icmlauthor{Bai Cong}{ttech}
\icmlauthor{Peter Nickl}{riken}
\icmlauthor{Gian Maria Marconi}{riken}
\icmlauthor{Clement Bazan}{ttech}
\icmlauthor{Rio Yokota}{ttech}
\icmlauthor{Iryna Gurevych}{tud}
\icmlauthor{Daniel Cremers}{tum}
\icmlauthor{Mohammad Emtiyaz Khan}{riken}
\icmlauthor{Thomas Möllenhoff}{riken}
\end{icmlauthorlist}

\icmlaffiliation{tum}{Technical University of Munich \& Munich Center
  for Machine Learning, Munich, Germany}
\icmlaffiliation{tud}{UKP Lab, Technical University of Darmstadt \& hessian.AI, Darmstadt, Germany}
\icmlaffiliation{ttech}{Tokyo Institute of Technology, Tokyo, Japan}
\icmlaffiliation{riken}{RIKEN Center for AI Project, Tokyo, Japan}

\icmlcorrespondingauthor{Thomas
  Möllenhoff}{thomas.moellenhoff@riken.jp}

\icmlkeywords{Bayesian deep learning, optimization, variational methods}

\vskip 0.3in
]



\printAffiliationsAndNotice{\icmlEqualContribution} 

\begin{abstract}
We give extensive empirical evidence against the common
belief that variational learning is ineffective for large neural networks. We
show that an optimizer called Improved
Variational Online Newton (IVON) consistently matches or outperforms
Adam for training large
networks such as GPT-2 and ResNets from scratch. 
IVON's computational costs are nearly
identical to Adam but its predictive uncertainty is better. 
We show several new use cases of IVON
where we improve finetuning and model merging in Large Language Models,
accurately predict generalization error,
and faithfully estimate sensitivity to data.
We find overwhelming evidence that variational
learning is effective.
   Code is available at \href{https://github.com/team-approx-bayes/ivon}{https://github.com/team-approx-bayes/ivon}.
\end{abstract}

\section{Introduction}
\label{sec:introduction}

Variational learning can potentially improve many aspects of deep learning, but there remain doubts about its effectiveness for large-scale problems. 
Popular strategies~\citep{Gr11, BlCo15} do not easily perform well, even on moderately-sized problems, 
which has led some to believe that it is impossible to get both good accuracy and uncertainty~\citep{TrTu17,FoBu20,CoBr22}. 
Variational methods generally have higher costs or tricky implementations~\citep{KiSa15,HeAd15,ZhSu18,KhNi18,OsSw19},
and they struggle to keep up with the ever-increasing scale of deep learning.

Currently, no variational method can accurately train Large Language Models (LLMs) from scratch at a cost, say, similar to Adam \citep{KiBa14}.~This is excluding methods such as MC-dropout~\citep{GaGh16}, stochastic weight averaging (SWAG)~\citep{MaIz19}, and Laplace~\citep{MK92}, 
which do not directly optimize the variational objective, even though they have variational interpretations.~Ideally, 
we want a direct optimization of the objective to match Adam's accuracy without increasing the cost, and also yield good weight-uncertainty
to improve calibration, model averaging, knowledge transfer, etc.

\begin{figure*}[t!]
  \centering
  \subfigure[GPT-2 on OpenWebText]{
      \includegraphics[width=.31\textwidth, keepaspectratio]{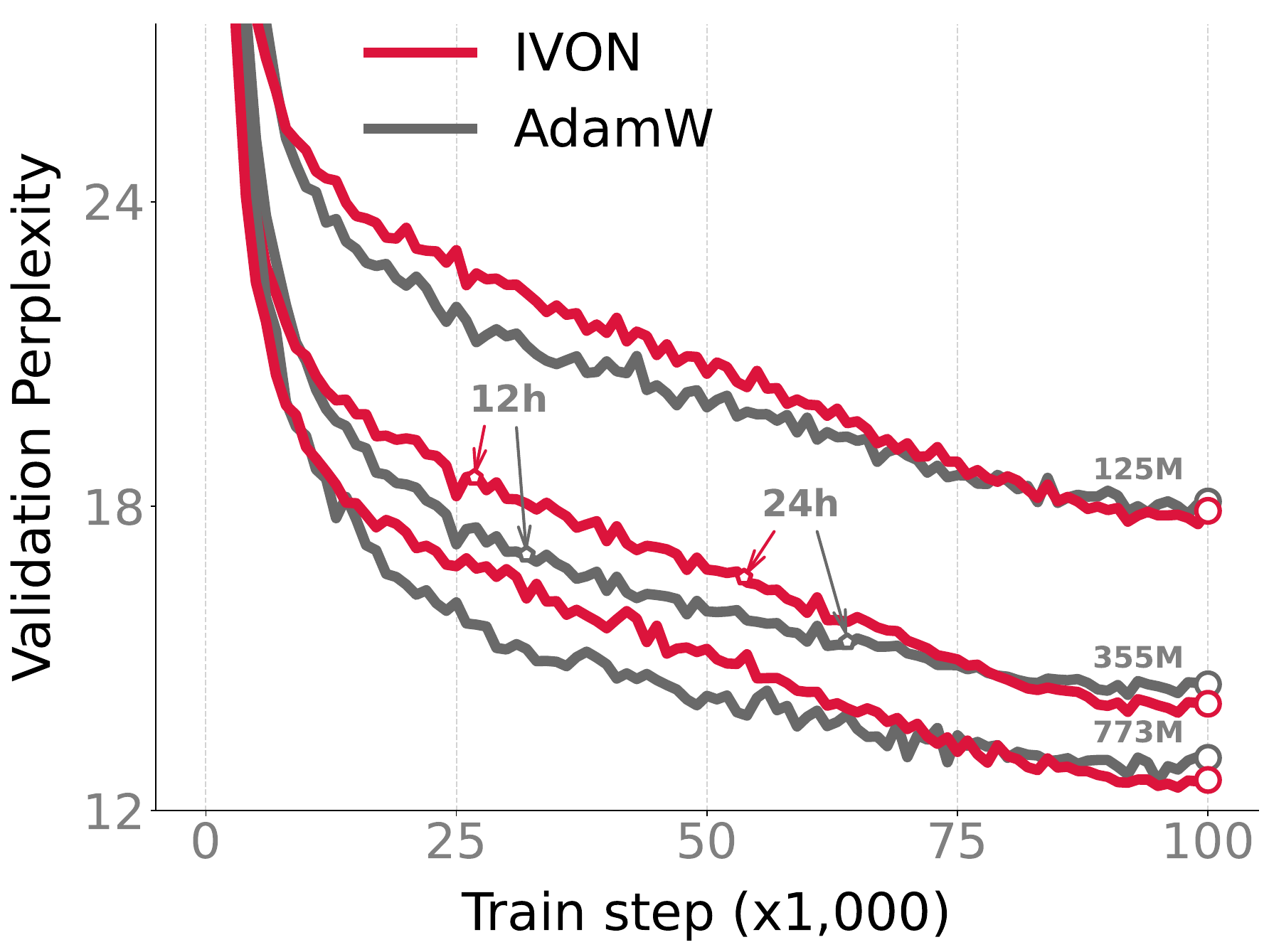}
    \label{fig:teaser_a}
  }
  \subfigure[ResNet-50 on ImageNet]{
    \includegraphics[width=.31\textwidth, keepaspectratio]{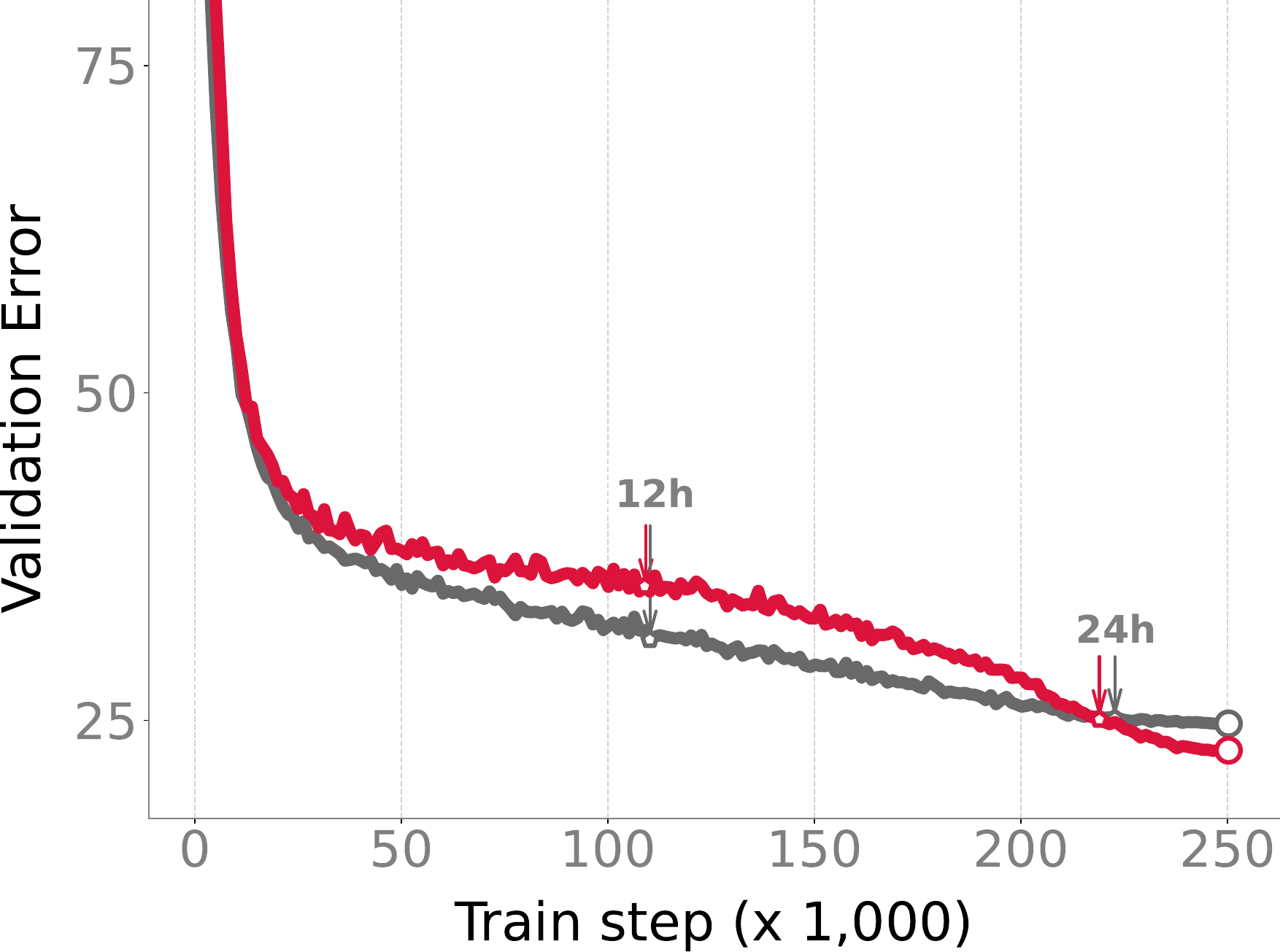}
    \label{fig:teaser_b}
  }
   \subfigure[Calibration on ImageNet]{
    \includegraphics[width=.31\textwidth, keepaspectratio]{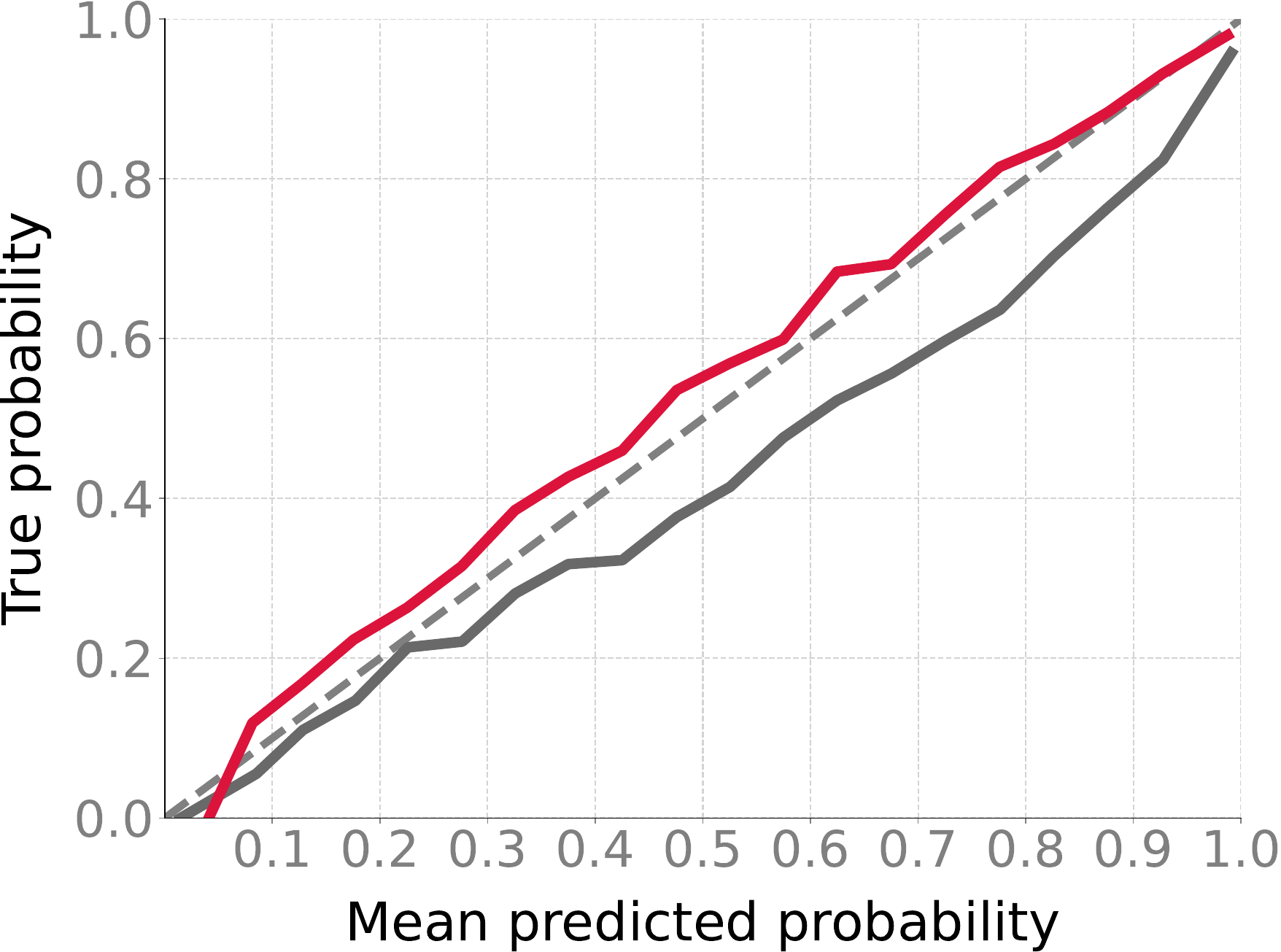}
    \label{fig:teaser_c}
  }
  \caption{First two panels show that IVON closely matches the
    trajectory of AdamW \citep{LoHu17} for training GPT-2 on
    OpenWebText and ResNet-50 on ImageNet. \revision{
    The computational costs of IVON and AdamW are nearly
    identical. Runtime in hours (h) is indicated by the arrows.}
  The third panel shows that the predictions are also better
  calibrated \revision{ as the red curve is closer to diagonal}. Comparisons to SGD on ImageNet are in
  \Cref{tab:imagenet}.
  Final numbers for IVON vs AdamW are as follows: 
  \revision{12.6} vs. 13.0  perplexity (lower is better) on GPT-2 (773M),
  14.1 vs 14.5
    perplexity on GPT-2 (355M), 17.9 vs 18.1
  perplexity on GPT-2 (125M), 77.5 vs 75.2 accuracy and 0.022 vs 0.066 ECE (lower is better) on ResNet-50.}
  \label{fig:teaser}
\end{figure*}

In this paper, we present the Improved Variational Online Newton (IVON) method, which adapts the method of~\citet{LiSc20} to large scale and obtains state-of-the-art accuracy and uncertainty at nearly identical cost as Adam.
\Cref{fig:teaser} shows some examples where, for training GPT-2 (773M parameters) from scratch, IVON gives \revision{0.4} reduction in validation perplexity over AdamW and, for ResNet-50 (25.6M parameters) on ImageNet, it gives around 2\% more accurate predictions that are also better calibrated. 
For image classification, we never observe severe overfitting like AdamW and consistently obtain better or comparable results to SGD. 

We introduce practical tricks necessary to achieve good performance and present an Adam-like implementation (\Cref{alg:ivon}) 
which uses a simplified Hessian-estimation scheme to both adapt the learning rate and estimate weight-uncertainty.~This also makes IVON a unique second-order 
optimizer that consistently performs better than Adam at a similar cost. We present extensive numerical experiments and new use cases to demonstrate its effectiveness. We find that,
\vspace{-6pt}
\begin{enumerate}[itemsep=-0.4pt]
   \item IVON gets better or comparable predictive uncertainty to alternatives, such as, MC-dropout and SWAG.
   \item It works well for finetuning LLMs and reduces the cost of model-merging.
   \item It can be used to faithfully predict generalization which is useful for diagnostics and early stopping.
   \item It is useful to understand sensitivity to data which is often challenging at large-scale due to ill-conditioning.
   \end{enumerate}
\vspace{-6pt}
Overall, we find overwhelming evidence that variational learning is not only effective but also useful for large deep networks, especially LLMs. 
IVON is easily amenable to flexible posterior forms \cite{LiKh19b}, and we expect it to help researchers further investigate the benefits of Bayesian principles to improve deep learning.

\section{Challenges of Variational Learning for Large Deep Networks}
Variational learning is challenging for large networks due to fundamental differences in its objective to those commonly used in deep learning. 
Deep learning methods estimate network weights $\vparam\in\real^P$ by minimizing empirical risk $\barloss(\vparam) = \sum_{i=1}^N \loss_i(\vparam)/N$, 
which is an average over individual losses $\loss_i(\vparam)$ for $N$ examples. In contrast, variational methods estimate a distribution $q(\vparam)$ over weights by minimizing 
\begin{equation}
   \cL(q) =  \lambda \bbE_{q (\vparam)} \left[  \barloss(\vparam) \right] + \myKL{q(\vparam)}{p(\vparam)}, \label{eq:ivonobjective}
\end{equation}
where $p(\vparam)$ is the prior, \revision{$\myKL{\cdot}{\cdot}$ the
  Kullback-Leibler divergence and $\lambda$ a scaling parameter often
set to $N$, but other values are useful, for example, to handle model misspecification.}
The objective in \Cref{eq:ivonobjective} coincides with variational \emph{inference} when
$\barloss(\vparam)$ is a proper likelihood. \revision{We use the term
  \emph{variational learning} to denote the general case}.

Optimization of $\cL(q)$ is fundamentally different from that of
$\barloss(\vparam)$. For instance, the number of parameters of $q$ can
be much larger than the size of $\vparam$, making the problem harder. The number of parameters of $q$ is doubled for a diagonal-covariance Gaussian $q(\vparam) = \gauss(\vparam\,|\,\vm,
\diag(\vsigma)^2)$ due to the estimation of two vectors of mean
$\vm\in\smash{\real^P}$ and standard deviation
$\vsigma\in\smash{\real^P}$, respectively. The optimization is further
complicated \revision{because of} the expectation in \cref{eq:ivonobjective}, which
adds additional noise during the optimization.

Due to these differences, a direct optimization of \cref{eq:ivonobjective} remains challenging. The standard approach is to optimize it by using a standard deep learning method, say, SGD,
\[
   \vm \leftarrow \vm - \rho \widehat{\nabla}_{\text{\vm}} \cL  \qquad  
      \vsigma \leftarrow \vsigma - \rho \widehat{\nabla}_{\text{\vsigma}} \cL,
\]
where $\rho>0$ is the learning rate. This showed promising results in
\revision{early attempts} of variational deep learning with several different stochastic
gradient estimators~$\widehat{\nabla}$~\citep{Gr11,BlCo15}.
Unfortunately, these methods have been unable to keep up with the growth in scale of deep learning.
The lack of progress has been attributed to various causes, such as high-variance in stochastic gradients~\citep{KiSa15,WeVi18}, issues with the temperature parameter~\citep{WeRo20,noci2021disentangling}, and lack of a good prior~\citep{FoGa22}. Multiple theoretical studies have raised doubts whether variational learning can ever work at all \citep{TrTu17,FoBu20,CoBr22}. 
Altogether, these have led to a belief \revision{that there exists} an inherent trade-off between accuracy and uncertainty in Bayesian learning.

Progress in variational learning has been made on a different front by
using \emph{natural-gradient} methods \cite{Sa01, HoBl13, KhLi17}
which have shown promising results on ImageNet \citep{OsSw19}. Their
updates resemble an Adam-like form which makes it easy to tune them at
large scale. Yet, the implementation can be tricky and the cost can be much higher than Adam. For example, \citet{OsSw19} build upon the Variational Online Newton (VON) method of
\citet{KhNi18} where they replace the Hessian computation by a Gauss-Newton estimate.
\revision{They implement the following Adam-like update:}
\begin{equation}
  \begin{split}
      \widehat{\vh} &\leftarrow \frac{1}{|\minibatch|} \sum_{i\in \minibatch} \nabla \loss_i(\vparam)^2, \quad \text{ where } \vparam \sim q,\\
      \vg &\leftarrow \beta_1 \vg + \widehat{\nabla} \barloss(\vparam)
      + s_0 \vm/\lambda,\\
      \vh &\leftarrow \beta_2 \vh + (1-\beta_2) \widehat{\vh}, \\
      \vm &\leftarrow \vm - \alpha_t \vg/(\vh + c), \\
      \vsigma &\leftarrow 1/\sqrt{\lambda(\vh+ c)}.
   \end{split}
   \label{eq:vogn}
 \end{equation}
\revision{Here, a prior $p(\vparam) =
\gauss(\vparam \, | \, 0, \vI/s_0)$ is used.}
The difficult computation is in the first line of~\Cref{eq:vogn} where a Gauss-Newton
estimate over a minibatch $\minibatch$ is computed at a sample from
the Gaussian, while the rest is similar to Adam:
\revision{the} second line is gradient momentum, where $s_0 \vm/\lambda$ is added due to the prior. 
\revision{The third and fourth line are identical to the scale and parameter vectors
updates, respectively.} The constant $c = \gamma+s_0/\lambda$ where $\gamma>0$ is \revision{a damping parameter}.   

The computation of the Gauss-Newton estimate is tricky because it requires per-example squaring, which is not a standard operation in deep learning and could be difficult to implement. In \citet[Fig. 1]{OsSw19}, this ends up increasing the cost by a factor of two. The Gauss-Newton estimate also introduces an additional approximation in the variational learning, even though it helps to ensure the positivity of $\vh$.
Another issue is the use of an additional damping parameter $\gamma$ which departs from the Bayesian framework.

Ideally, we want a method that directly optimizes \cref{eq:ivonobjective} without additional approximations and also seamlessly fits into an Adam-like framework without any significant computational overheads. 
Methods such as MC-dropout, SWAG, and Laplace do not solve this problem, and rather circumvent it by relying on algorithms that optimize $\barloss$, not $\cL$. The goal of
this paper is to propose a method that can match the accuracy of Adam while directly optimizing $\cL$.

\section{Improved Variational Online Newton}
\label{sec:ivon}

We present the Improved Variational Online Newton (IVON) method by adapting the method of~\citet{LiSc20} and introducing practical tricks necessary to achieve good performance at large scale. 
They propose an improved version of \revision{the} Bayesian Learning Rule \citep{KhRu21} which ensures positivity of certain variational parameters, such as, 
the Gaussian variance or scale parameter of a Gamma distribution. For the Gaussian case, they propose an Adam-like update which makes the update in \cref{eq:vogn}
simpler. Specifically, they use the following Hessian estimate by using the reparameterization trick,
\begin{equation}
   \widehat\vh \leftarrow \widehat{\nabla} \barloss(\vparam) \cdot \frac{\vparam - \vm}{\vsigma^2} , \label{eq:lh_estimator}
\end{equation}
which does not require per-example gradient squares, rather just a
single vector multiplication with the minibatch gradient. 
The above estimate is easy to compute but, unlike the Gauss-Newton estimate, it
is not always positive and can make $\vh$ in \cref{eq:vogn} negative
\citep[App. D]{KhNi18}. \citet{LiSc20} solve this problem by using Riemannian gradient descent which ensures positivity by adding an extra term in the update of $\vh$,
\begin{equation}
   \vh \leftarrow (1 - \rho) \vh + \rho \widehat{\vh} + \half \rho^2
   (\vh - \widehat{\vh})^2 / (\vh + s_0 / \lambda), \label{eq:update-dot-improved}
\end{equation}
where $\rho>0$ is a constant.
Positivity holds even when $\widehat{\vh}$ are negative, \revision{as shown in \citet[Theorem~1]{LiSc20}.}

In \cref{alg:ivon}, we use the two modifications (highlighted in red) to get an improved version of VON, called IVON. The updates closely resemble Adam, but there is a sampling step in line 2 (highlighted in blue) and there is no square-root over $\vh$ in line 7. IVON therefore uses a Newton-like update.
The Hessian estimator in \cref{eq:lh_estimator} is less costly compared to other second-order optimizers~\citep{DaDe15,YaGh21,LiLi23}. It is valid even for losses that are not twice-differentiable (for example, for ReLU activations). These aspects make IVON a unique second-order optimizer with similar costs to Adam.
\definecolor{commentcolor}{RGB}{128, 179, 89}
\renewcommand\algorithmiccomment[1]{\hfill{\textcolor{commentcolor}{\eqparbox{COMMENT}{#1}}}}
\begin{algorithm}[t!]
   \caption{Improved Variational Online Newton (IVON). Hyperparameter
     setting is described in~\Cref{subsec:guide}.}
	\label{alg:ivon}
	\begin{algorithmic}[1]
		\setstretch{1.15}
      \REQUIRE Learning rates $\{ \alpha_t \}$, weight-decay $\delta >
      0$.
      \REQUIRE Momentum parameters $\beta_1, \beta_2 \in [0, 1)$.
      \REQUIRE Hessian init $h_0 > 0$.
      \renewcommand{\algorithmicrequire}{\textbf{Init:}}
		\REQUIRE $\vm \leftarrow \text{(NN-weights)}$,\,\, $\vh \leftarrow h_0 $,
                      \,\, $\vg \leftarrow 0$, \,\, $\lambda \leftarrow N$.
\REQUIRE $\vsigma
                      \leftarrow 1 / \sqrt{\lambda (\vh +
                        \delta)}$.
      \renewcommand{\algorithmicrequire}{\textbf{Optional:}}
\REQUIRE $\alpha_t \leftarrow (h_0 +
                      \delta) \alpha_t$\,  for all $t$.
                      \FOR{$t=1,2,\hdots$}
                      \STATE \hspace{-0.15cm}$\widehat \vg \leftarrow
                      {\widehat \nabla} \barloss(\vparam)$,
      \text{\textcolor{black}{ where} } 
      $\color{blue}\vparam \sim q$
      \STATE \hspace{-0.15cm}$\widehat \vh \leftarrow {\color{crimson} \widehat \vg\cdot (\vparam-\vm) / \vsigma^2}$
		\STATE \hspace{-0.15cm}$\vg \leftarrow \beta_1 \vg\hspace{-0.03cm}+\hspace{-0.05cm}(1\hspace{-0.05cm}-\hspace{-0.05cm}\beta_1) \widehat \vg$ 
                \STATE \hspace{-0.15cm}$\vh \leftarrow \beta_2 \vh+(1
                - \beta_2)\widehat \vh$${\color{crimson}+\half (1 - \beta_2)^2
                  (\vh - \widehat \vh)^2 / (\vh + \delta)}$
      \STATE \hspace{-0.15cm}$\bar \vg \leftarrow \vg / (1 - \beta_1^{t})$ 
      \STATE \hspace{-0.15cm}$\vm \leftarrow \vm - \alpha_t(\bar \vg + \delta \vm) / ({\color{blue} \vh} + \delta)$
		\STATE \hspace{-0.15cm}$\vsigma \leftarrow 1 / \sqrt{\lambda (\vh + \delta)}$
      \ENDFOR
		\STATE \textbf{return} $\vm, \vsigma$ 
	\end{algorithmic}
	\setstretch{1}
      \end{algorithm}

            \begin{table*}[t!]
                \small
                \centering
            \begin{tabular}{llrccccc}
                \toprule
                \,\,\, Dataset / Model & Method & & Top-1 Acc. $\uparrow$ & Top-5 Acc. $\uparrow$ & NLL $\downarrow$ & ECE $\downarrow$ & Brier $\downarrow$ \\
            \midrule
              & AdamW && \hspace{-0.95cm}\makebox[0.85cm][r]{\scriptsize \textcolor{crimson}{($2\%$)}} $75.16_{\pm 0.14}$ & $92.37_{\pm 0.03}$ & $1.018_{\pm 0.003}$ & $0.066_{\pm 0.002}$ & $0.349_{\pm 0.002}$ \\
              & SGD &&\hspace{-0.95cm}\makebox[0.85cm][r]{\scriptsize \textcolor{crimson}{($1\%$)}} $76.63_{\pm 0.45}$ & $93.21_{\pm 0.25}$ & $0.917_{\pm 0.026}$ & $0.038_{\pm 0.009}$ & $0.326_{\pm 0.006}$ \\
             \rowcolor{gray!10} \cellcolor{white} & IVON@mean && $77.30_{\pm 0.08}$ & $93.58_{\pm 0.05}$ & $0.884_{\pm 0.002}$ & $0.035_{\pm 0.002}$ & ${\bf 0.316}_{\pm 0.001}$ \\
             \rowcolor{gray!10}  \cellcolor{white}\multirow{-4}*{\begin{tabular}{l}\cellcolor{white}\text{ImageNet} \\\cellcolor{white}\text{ResNet-50}\\\cellcolor{white}{(26M params)}  \end{tabular}} & 
              IVON && ${\bf 77.46}_{\pm 0.07}$ & ${\bf 93.68}_{\pm 0.04}$ & ${\bf 0.869}_{\pm 0.002}$ & ${\bf 0.022}_{\pm 0.002}$ & ${\bf 0.315}_{\pm 0.001}$ \\
            \midrule
                 & AdamW && \hspace{-1cm}\makebox[0.85cm][r]{\scriptsize \textcolor{crimson}{($15\%$)}} $47.33_{\pm 0.90}$ & $71.54_{\pm 0.95}$ & $6.823_{\pm 0.235}$ & $0.421_{\pm 0.008}$ & $0.913_{\pm 0.018}$ \\
                 & SGD && \hspace{-1cm}\makebox[0.85cm][r]{\scriptsize \textcolor{crimson}{($1\%$)}} $61.39_{\pm 0.18}$ & $82.30_{\pm 0.22}$ & $1.811_{\pm 0.010}$ & $0.138_{\pm 0.002}$ & $0.536_{\pm 0.002}$ \\
                \rowcolor{gray!10} \cellcolor{white} & IVON@mean && ${\bf 62.41}_{\pm 0.15}$ & ${\bf 83.77}_{\pm 0.18}$ & $1.776_{\pm 0.018}$ & $0.150_{\pm 0.005}$ & $0.532_{\pm 0.002}$ \\
                \rowcolor{gray!10}  \cellcolor{white}\multirow{-4}*{\begin{tabular}{l}\cellcolor{white}\text{TinyImageNet}\\\cellcolor{white}\text{ResNet-18}\\\cellcolor{white}(11M params)\\ \end{tabular}} & 
             IVON && ${\bf 62.68}_{\pm 0.16}$ & ${\bf 84.12}_{\pm 0.24}$ & ${\bf 1.528}_{\pm 0.010}$ & ${\bf 0.019}_{\pm 0.004}$ & ${\bf 0.491}_{\pm 0.001}$ \\
                 \midrule
              & AdamW && \hspace{-1cm}\makebox[0.85cm][r]{\scriptsize
                         \textcolor{crimson}{($11\%$)}} $\smash{50.65_{\pm 0.0^{\ast}}}$ & $\smash{74.94_{\pm 0.0^{\ast}}}$ & $\smash{4.487_{\pm 0.0^{\ast}}}$ & $\smash{0.357_{\pm 0.0^{\ast}}}$ & $\smash{0.812_{\pm 0.0^{\ast}}}$ \\
              & SGD && \hspace{-1cm}\makebox[0.85cm][r]{\scriptsize \textcolor{crimson}{($2\%$)}} $59.39_{\pm 0.50}$ & $81.34_{\pm 0.30}$ & $2.040_{\pm 0.040}$ & $0.176_{\pm 0.006}$ & $0.577_{\pm 0.007}$ \\
             \rowcolor{gray!10} \cellcolor{white} & IVON@mean && ${\bf 60.85}_{\pm 0.39}$ & ${\bf 83.89}_{\pm 0.14}$ & $1.584_{\pm 0.009}$ & $0.053_{\pm 0.002}$ & ${\bf 0.514}_{\pm 0.003}$ \\
             \rowcolor{gray!10}  \cellcolor{white}\multirow{-4}*{\begin{tabular}{l}\cellcolor{white}\text{TinyImageNet}\\\cellcolor{white}\text{PreResNet-110}\\\cellcolor{white}(4M params)\\ \end{tabular}} & 
             IVON && ${\bf 61.25}_{\pm 0.48}$ & ${\bf 84.13}_{\pm 0.17}$ & ${\bf 1.550}_{\pm 0.009}$ & ${\bf 0.049}_{\pm 0.002}$ & ${\bf 0.511}_{\pm 0.003}$ \\
             \midrule
                  & AdamW & &
                              \hspace{-1cm}\makebox[0.85cm][r]{\scriptsize \textcolor{crimson}{($11\%$)}} $64.12_{\pm 0.43}$ & $86.85_{\pm 0.51}$ & $3.357_{\pm 0.071}$ & $0.278_{\pm 0.005}$ & $0.615_{\pm 0.008}$ \\
                  & SGD && \hspace{-1cm}\makebox[0.85cm][r]{\scriptsize
                           \textcolor{crimson}{($1\%$)}} $74.46_{\pm 0.17}$ & $92.66_{\pm 0.06}$ & $1.083_{\pm 0.007}$ & $0.113_{\pm 0.001}$ & $0.376_{\pm 0.001}$ \\
                 \rowcolor{gray!10}  \cellcolor{white} & IVON@mean && $74.51_{\pm 0.24}$ & $92.74_{\pm 0.19}$ & $1.284_{\pm 0.013}$ & $0.152_{\pm 0.003}$ & $0.399_{\pm 0.002}$ \\
                 \rowcolor{gray!10}
              \cellcolor{white}\multirow{-4}*{\begin{tabular}{l}\cellcolor{white}\text{CIFAR-100}\\\cellcolor{white}\text{ResNet-18}\\\cellcolor{white}(11M
                                                params)\\\end{tabular}} &
                                                                          IVON && ${\bf 75.14}_{\pm 0.34}$ & ${\bf 93.30}_{\pm 0.19}$ & ${\bf 0.912}_{\pm 0.009}$ & ${\bf 0.021}_{\pm 0.003}$ & ${\bf 0.344}_{\pm 0.003}$ \\
             \midrule
                                       & AdamW && \hspace{-1cm}\makebox[0.85cm][r]{\scriptsize
                         \textcolor{crimson}{($10\%$)}} $65.88_{\pm 0.84}$ & $88.34_{\pm 0.56}$ & $2.893_{\pm 0.088}$ & $0.258_{\pm 0.006}$ & $0.578_{\pm 0.014}$ \\
                                       & SGD && \hspace{-1cm}\makebox[0.85cm][r]{\scriptsize
                         \textcolor{crimson}{($2\%$)}} $74.19_{\pm 0.11}$ & $92.41_{\pm 0.14}$ & $1.204_{\pm 0.012}$ & $0.137_{\pm 0.002}$ & $0.393_{\pm 0.004}$ \\
              \rowcolor{gray!10} \cellcolor{white}& IVON@mean && $75.23_{\pm 0.23}$ & $93.45_{\pm 0.16}$ & $1.149_{\pm 0.010}$ & $0.136_{\pm 0.002}$ & $0.380_{\pm 0.003}$ \\
              \rowcolor{gray!10} \cellcolor{white}\multirow{-4}*{\begin{tabular}{l}\cellcolor{white}CIFAR-100\\\cellcolor{white}PreResNet-110\\ \cellcolor{white}(4M params) \end{tabular}} & IVON && ${\bf 75.81}_{\pm 0.18}$ & ${\bf 93.93}_{\pm 0.19}$ & ${\bf 0.884}_{\pm 0.007}$ & ${\bf 0.030}_{\pm 0.003}$ & ${\bf 0.336}_{\pm 0.001}$ \\
            \bottomrule
        \end{tabular}
      
        \caption{IVON improves both accuracy and uncertainty over SGD
          and AdamW. \revision{Improvements in accuracy by IVON are shown in red. The performance of AdamW is not good on the smaller datasets likely due to overfitting when training for $200$ epochs. IVON does not have this issue.} Additional results are in the appendix in~\Cref{tab:resnets:cifar10,tab:resnets:cifar100,tab:resnets:tinyimagenet} and \Cref{tab:rnn_classification}. \vspace{-0.2cm} \label{tab:imagenet}}
    \end{table*}

Below are a few practical tricks needed for good results.
\vspace{-6pt}
\begin{enumerate}
   \item Instead of the prior precision $s_0$, we use the weight-decay
     regularizer $\delta$ as the prior. The scaling parameter
     $\lambda$ is set to $N$, except for finetuning on small
     datasets.
   \item  Unlike \citet[Fig.~1]{LiSc20}, the update of $\vh$ does not
     use $\delta$. We do not debias $\vh$ and we update it before $\vm$ which has no impact on the performance.
   \item The Hessian $\vh$ is initialized with a constant $h_0$. \citet{LiSc20} most likely set it to 0 due to the debiasing step used in their work. We find the initialization to be useful. Too small values can destabilize the training while larger values may give poor performance. 
       \item When training transformers, it can be helpful to clip the preconditioned
         gradient in~line~7 entrywise to $[-\xi, \xi]$.
   \item Optionally, we rescale $\alpha_t$ by $(h_0 + \delta)$ so that
       the first steps of the algorithm have step-size close to the
       initial $\alpha_t$. When clipping is used, this step
       is omitted.
\end{enumerate}
Momentum $\beta_1$, learning rate
$\alpha_t$ and weight-decay $\delta$ can be
set in the same fashion as for standard optimizers, as well as
minibatch size and \revision{clipping radius $\xi$}. $\beta_2$ typically needs to be
closer to one as in Adam, for instance, values of $\beta_2 = 0.99995$ work well.
$h_0$ is typically around $0.01$ to $1$, and setting $\lambda$ is also easy, as discussed above. 
This makes obtaining good results with IVON often very easy. A detailed guide for hyperparameter
setting is in~\Cref{subsec:guide}.

IVON can be easily modified to accommodate multi-GPU training. Each GPU device can use different Monte-Carlo (MC) samples, which reduces the variance \citep{KiSa15}. Moreover, multiple MC samples per device can also be used.~These modifications can be realized by simply replacing the calculations of $\widehat{\vg}$ and $\widehat{\vh}$ in line~2 and~3 of \Cref{alg:ivon} by the following averages over $J$ devices and a total of $S$ MC samples drawn on each device:
  \begin{align*}
    \widehat \vg = \frac{\sum_{j,s} \widehat \nabla \bar
    	\ell(\vparam_j^{(s)})}{J \cdot S}, ~~ \widehat \vh = \frac{\sum_{j,s} \widehat \nabla \bar \ell(\vparam_j^{(s)}) (\vparam_j^{(s)} - \vm)}{J \cdot S \cdot \vsigma^2} .
  \end{align*}
Here, we use a different $\vparam_j^{(s)} \sim q$ on each device $j$ and for each MC sample $s$. Both sums can be implemented as running averages for better memory-efficiency.

\revision{We implement IVON as
	a drop-in replacement for Adam in PyTorch, 
	where only two lines need to be
	added (marked red below) to draw multiple MC samples:}

\begin{minipage}{0.45\linewidth}
	\small
	\begin{Verbatim}[commandchars=\\\{\}]
for inputs, targets in dataloader:
 \textcolor{crimson}{for _ in range(num_mc_samples):}
  \textcolor{crimson}{with optimizer.sampled_params(train=True):}
   optimizer.zero_grad()
   outputs = model(inputs)
   loss = loss_fn(outputs, targets)
   loss.backward()
 optimizer.step()
	\end{Verbatim}
\end{minipage}

The first red line is often not needed because a single sample can already give reasonable results.

\begin{figure*}[t!]
	\centering
        \subfigure[Training loss on OpenWebText]{
          \includegraphics[height=1.57in]{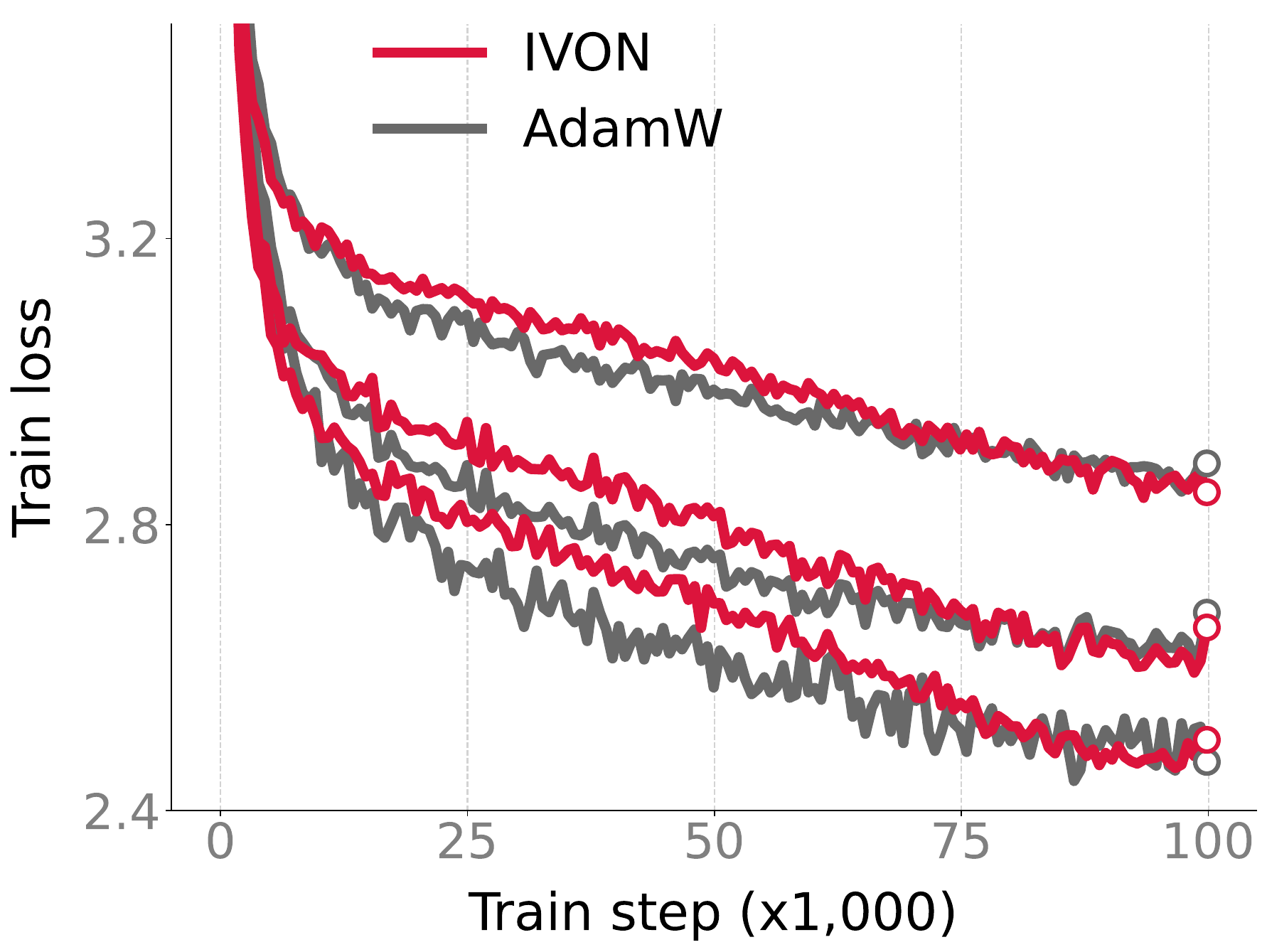}
          \label{fig:gpt2_train}
	}
        \subfigure[Low precision training]{
            \includegraphics[height=1.57in]{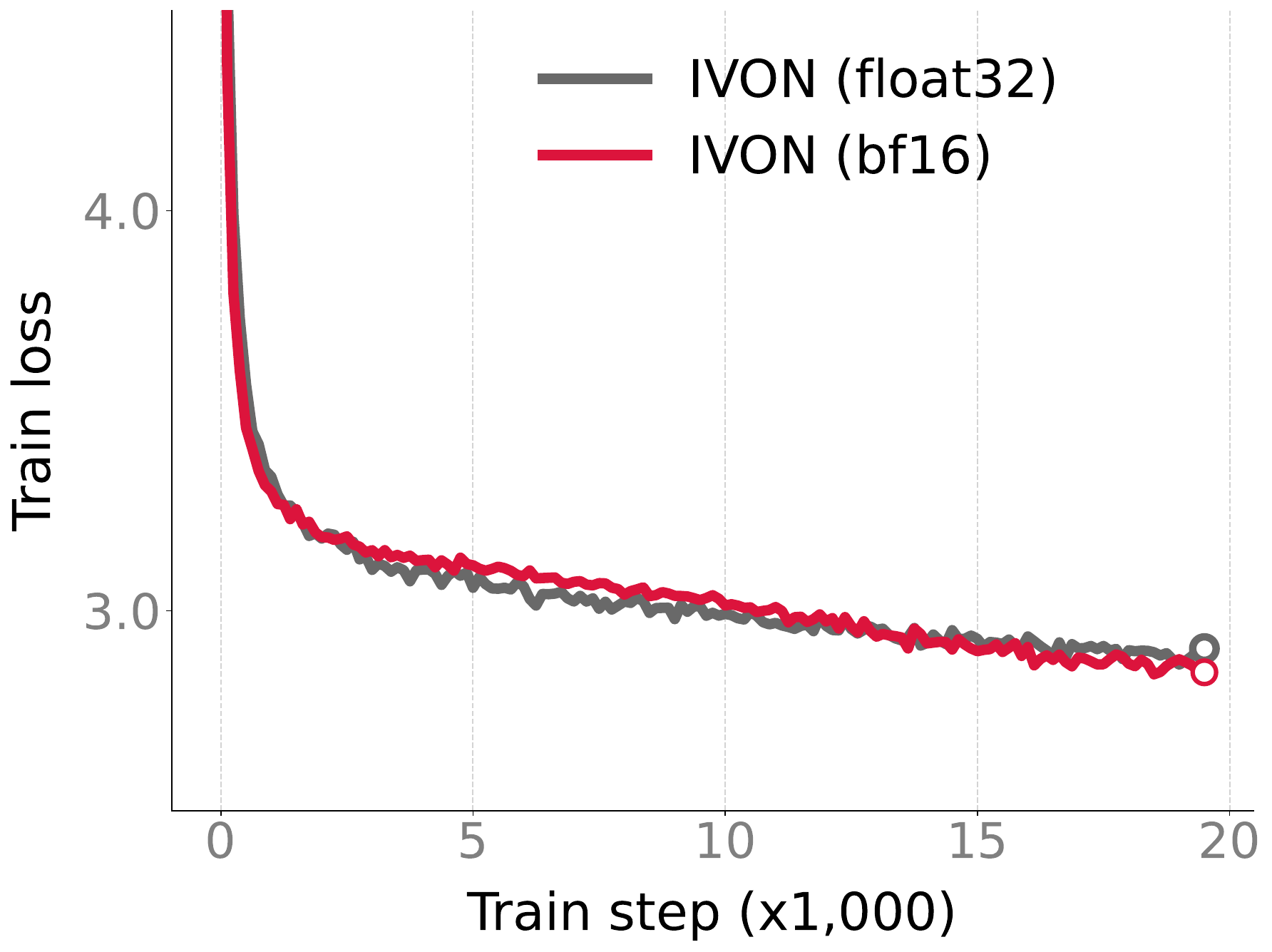}
        \label{fig:bf16gpt2}
      }
	\subfigure[Predictive posterior]{
        \includegraphics[height=1.57in]{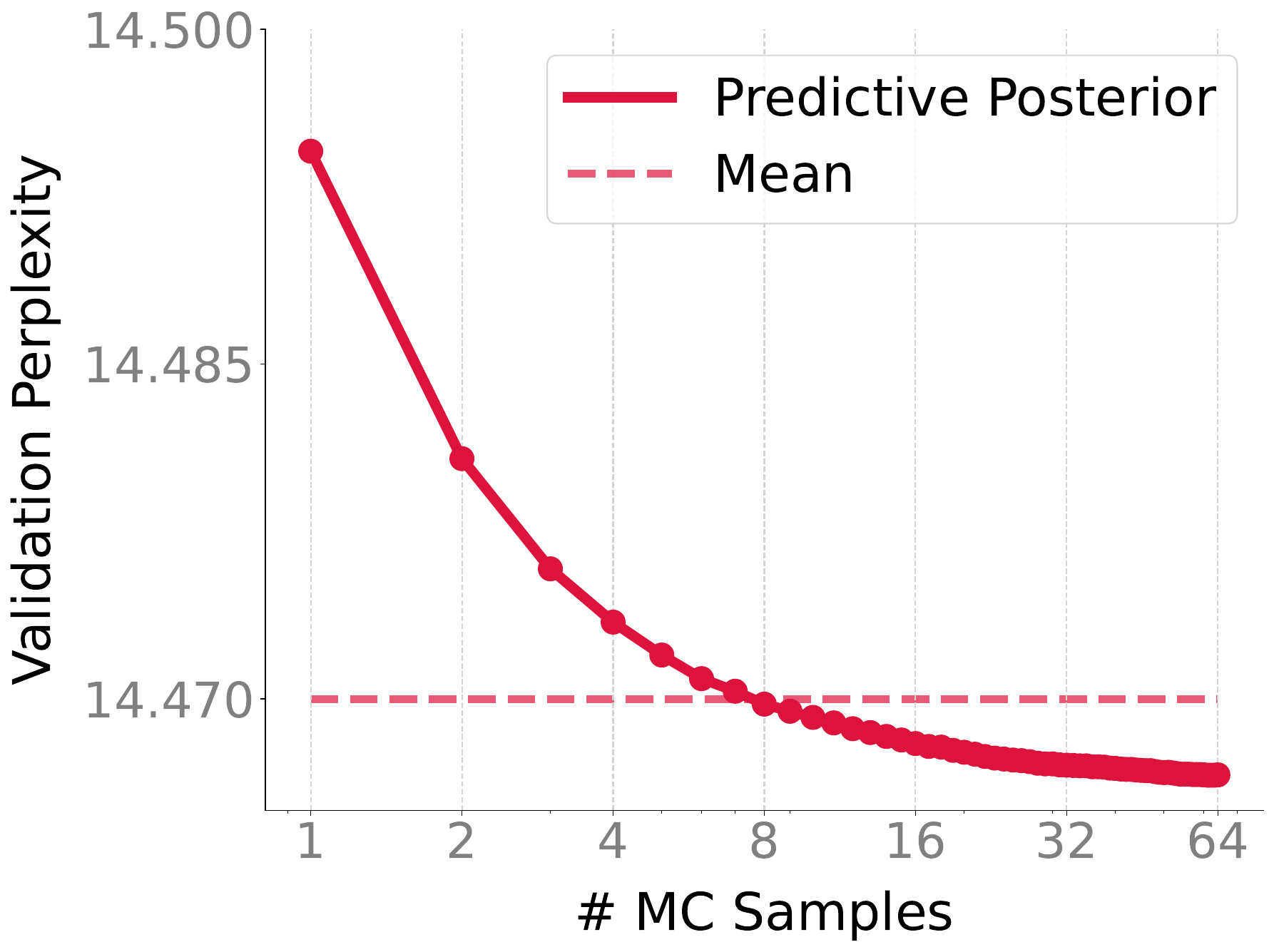}
	\label{fig:mcsamplesgpt2}
        }
	\caption{\revision{Panel (a) shows that, when training GPT-2, IVON not only improves upon AdamW in terms of
          validation perplexity but also converges to matching or even
          better training loss than AdamW. Panel b shows that IVON
          provides stable training when using low-precision (bf16)
          floating point numbers. Panel (c) shows that averaging predictions over IVON's posterior further improves the validation perplexity on GPT-2, when a sufficient number of samples is used ($>$ 8).}   }
\end{figure*}

\section{IVON is Effective for Large Deep Networks}
\label{sec:secondorder}
\revision{We show that IVON effectively trains large deep networks from scratch (\Cref{subsec:benefit-generalization}) and enables many 
downstream applications, such as predictive uncertainty (\Cref{subsec:benefit-uncertainty}), finetuning and model merging (\Cref{subsec:benefit-model-merging}), 
as well as predicting generalization and 
finding influential training examples (\Cref{subsec:benefit-data-sensitivity}).}
\revision{We perform ablation studies
  on computational
  efficiency~(\Cref{app:efficiency}), the choice of Hessian
  estimator~(\Cref{app:exp-ablation-hess}) and initialization~(\Cref{app:hess-init-ablation}).}
In the following, we refer by IVON@mean to the prediction using $\vm$ as the weights, whereas IVON denotes a model average
with $64$ samples drawn from the posterior learned by IVON.

\subsection{Better Scalability and Generalization}
\label{subsec:benefit-generalization}
\revision{Here, we show how IVON scalably trains large models from scratch.
First, we train LLMs with up to $773$M parameters from scratch on ca. $50$B tokens. 
Then, we show improved accuracy and uncertainty for various
image classification models, for example, ResNets with 26M parameters at
ImageNet-scale.
Additional results on smaller recurrent neural networks with 2M
parameters are in~\Cref{app:rnn}.
}

\subsubsection{Pretraining language models}
\label{sec:gpt_training}
\revision{
  Pretraining transformer language models~\citep{VaSh17}
with variational learning has been challenging and no large-scale result exists so far.
We show that IVON can train large language models at scale.}
We train models following the GPT-2 architecture~\citep{RaWu2019} for $49.2$ billion tokens in total on the OpenWebText corpus~\citep{GoCo19}.
We use the same hyperparameters for AdamW as prior work~\citep{LiLi23}. For IVON, we set them by grid search on a smaller model.
We pretrain from scratch three models with parameter sizes of $125$M, $355$M (``GPT-2-medium''), and $773$M (``GPT-2-large''), respectively. We use gradient clipping to stabilize the training.
Details are in~\Cref{app:gpt2_details}.

As shown in~\Cref{fig:teaser_a}, for the three models, the validation perplexities are reduced from $18.1$ to $17.9$, from
$14.5$ to $14.1$ and from $13.0$ to \revision{$12.6$}, respectively.
\Cref{fig:gpt2_train} further shows the same trend for training loss.
\Cref{fig:bf16gpt2} shows stable training with bf16 precision, and
\Cref{fig:mcsamplesgpt2} shows
that sampling multiple models from IVON's posterior further improves performance \revision{when a sufficient number of samples is used.}
\revision{Overall, we see effectiveness for training large Transformers from scratch on large datasets.}

\subsubsection{Image classification}
\label{subsec:imagenet}
\revision{We compare IVON to AdamW and SGD\revision{ (with momentum)} for image classification on various models and benchmarks.}
\Cref{tab:imagenet} shows that IVON improves upon both AdamW and the stronger SGD baseline in terms of both accuracy and uncertainty, here measured by 
negative log-likelihood (NLL), expected calibration error (ECE), and
Brier score. We also find that IVON does not overfit on smaller tasks,
unlike AdamW which tends to overfit on TinyImageNet and CIFAR-100.
\revision{This holds on various datasets and models trained for $200$ epochs. We show two of them here.
First, we show ResNet-50 with around $25.6$M
parameters~\citep{HeZh16} on ImageNet-1k which has around $1.2$M images with $1000$ classes. 
Second, we show ResNet-18 with 11M parameters and PreResNet-110 with $4$M parameters on both TinyImageNet and CIFAR-100.
We list further details on the experiments in~\Cref{app:hyp2} along
with more results using DenseNet-121 and ResNet-20 on other datasets, such as CIFAR-10. There too, we find improvements in both accuracy and uncertainty.
}

\begin{figure*}[t!] 
    \vspace{-.15in}
	\centering
	\subfigure[CIFAR-10 on SVHN]{
		\includegraphics[width=.33\linewidth]{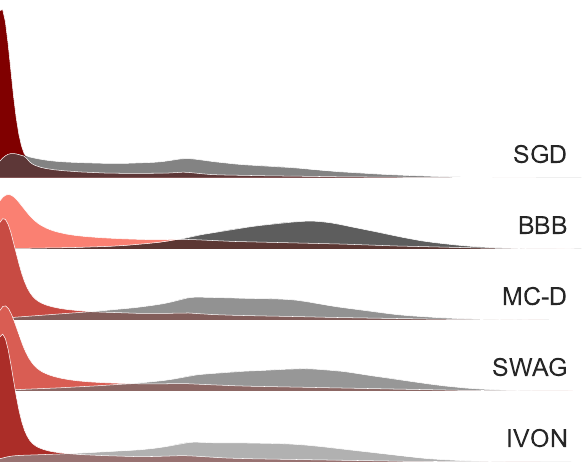}
        \label{fig:flat_c}}
        \subfigure[CIFAR-10 on Flowers102]{
		\includegraphics[width=.33\linewidth]{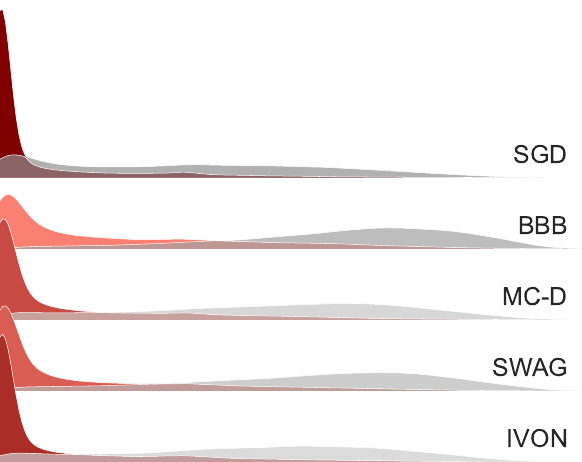}
		\label{fig:flat_d}
              }
              \subfigure[IVON in-between uncertainty]{\includegraphics[width=0.29\linewidth]{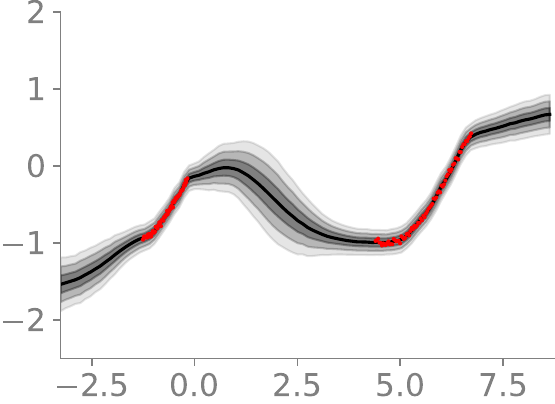}
              \label{fig:in-between}}
              \caption{\revision{In panel (a) and (b), we see that IVON's
                histogram of predictive entropy has a high peak
                similar to SGD for in-domain data (red, CIFAR-10) but
                at the same time is spread out widely similar
                to the other Bayesian deep learning methods for
                out-of-domain data (gray).
                The colors are shaded proportional to the height of
                the peak, that is, darker red and gray indicates
                a higher peak. In panel (c), we see that IVON can handle
                in-between uncertainty well, which has been shown to
                be challenging for variational methods by~\citet{foong2019between}.}}
	\label{fig:ood}
    \end{figure*}

\revision{
  We hypothesize that these improvements are partly due to flat-minima seeking
  properties of variational learning. Methods aimed to find flat
  minima, such as sharpness-aware
  minimization (SAM)~\citep{FoKl21}, have recently gained in popularity to
  boost test-accuracy.
\citet{MoKh23} have shown that SAM optimizes a relaxation of the expected 
loss in~\Cref{eq:ivonobjective}. Our results here indicate that
 similar improvements can be obtained by directly optimizing the variational objective and without using any relaxations.
}

\subsection{Posterior Averaging for Predictive Uncertainty}
\label{subsec:benefit-uncertainty}
\revision{Variational learning naturally allows for improved predictive uncertainties, \revision{because we can average predictions of several $\vparam$ sampled from the posterior.}
Unlike other Bayesian Deep Learning (BDL) methods, no postprocessing or model architecture changes are required for this.}
\revision{In the following, we compare IVON to AdamW, SGD,}
Bayes-by-Backprop \revision{(BBB)}, linearized last-layer Laplace (referred to as Laplace), MC Dropout (referred to as MC-D), SWAG\revision{, VOGN} and deep
ensembles~\citep{LaPrBl17}.~We consider both in-domain and out-of-domain (OOD) settings. We report common metrics from existing benchmarks~\citep{LiLi18, SnOv19}.
Further details on the experimental setup are in~\Cref{app:hyp-bdl}.

\begin{table}[t!]
\centering
\setlength{\tabcolsep}{3pt}
\resizebox{\linewidth}{!}{\begin{tabular}{lcccc}
\toprule
 & Acc. (\%) $\uparrow$ & NLL $\downarrow$ & ECE $\downarrow$ & Brier $\downarrow$ \\
\midrule
AdamW & $90.04_{\pm 0.27}$ & $0.589_{\pm 0.018}$ & $0.074_{\pm 0.002}$ & $0.170_{\pm 0.004}$ \\
SGD         & $91.86_{\pm 0.14}$ & $0.288_{\pm 0.015}$ & $0.040_{\pm 0.004}$ & $0.126_{\pm 0.004}$ \\
BBB         & $91.09_{\pm 0.16}$ & $0.289_{\pm 0.005}$ & $0.053_{\pm 0.001}$ & $0.139_{\pm 0.002}$ \\
Laplace	& $91.52_{\pm 0.37}$ & $0.284_{\pm 0.008}$ & $0.033_{\pm 0.002}$ & $0.129_{\pm 0.004}$ \\
MC-D    	& $91.85_{\pm 0.17}$ & $0.242_{\pm 0.004}$ & ${\bf 0.008}_{\pm 0.002}$ & $0.120_{\pm 0.002}$ \\
SWAG        & $92.45_{\pm 0.23}$ & $0.230_{\pm 0.002}$ & $0.024_{\pm 0.002}$ & $0.112_{\pm 0.002}$ \\
VOGN		& $92.37_{\pm 0.23}$ & $0.226_{\pm 0.005}$ & ${\bf 0.008}_{\pm 0.001}$ & $0.111_{\pm 0.003}$ \\ 
\rowcolor{gray!10} IVON & ${\bf 92.71}_{\pm 0.07}$ & ${\bf 0.219}_{\pm 0.002}$ & ${\bf 0.008}_{\pm 0.001}$ & ${\bf 0.108}_{\pm 0.001}$ \\
\cmidrule{1-5}
Deep Ens. & $93.57_{\pm 0.16}$ & $0.198_{\pm 0.003}$ & ${\bf 0.014}_{\pm 0.001}$ & $0.096_{\pm 0.001}$ \\
\rowcolor{gray!10} Multi-IVON    & ${\bf 94.37}_{\pm 0.13}$ & ${\bf 0.179}_{\pm 0.002}$ & $0.029_{\pm 0.001}$ & ${\bf 0.087}_{\pm 0.001}$ \\
\bottomrule
\end{tabular}}
\caption{IVON's predictive uncertainty is better than other baselines
  for in-domain examples. 
 Multi-IVON is a mixture-of-Gaussian ensemble which further improves
 the performance and is competitive with a deep ensemble.}
\label{tab:bdl-indomain}
\end{table}

\subsubsection{In-domain comparisons} 
To evaluate in-domain uncertainty, we train and evaluate ResNet-20 models~\citep{HeZh16} on the smaller CIFAR-10 dataset. 
\revision{We choose smaller datasets because it is difficult to apply BBB on larger problems.}
Results are reported in Table~\ref{tab:bdl-indomain}.
Overall, all BDL baselines, except for BBB which is known to underperform, 
have significantly better performance than SGD and AdamW. 
Among all non-ensemble approaches, IVON stands out in both accuracy and uncertainty metrics.

Deep ensembles made up of five models from different SGD runs clearly improve over the non-ensemble methods. We compare them to a similar version of IVON with a mixture-of-Gaussian posterior,
constructed from five independently-trained IVON models. 
This is referred to as Multi-IVON in Table~\ref{tab:bdl-indomain},
\revision{where we find it to outperform deep ensembles.} 
These results altogether confirm the quality of uncertainty estimates 
obtained with IVON.

    \begin{table}
        \setlength{\tabcolsep}{2pt}
        \centering
        \resizebox{\linewidth}{!}{
            \begin{tabular}{lccccc}
                \toprule
                & FPR@95\% $\downarrow$ & Det. Err. $\downarrow$ & AUROC $\uparrow$ & AUPR-In $\uparrow$ & AUPR-Out $\uparrow$ \\
              \midrule
              \multicolumn{6}{c}{\textbf{SVHN}, see~\Cref{fig:flat_c}}\\
              \midrule
SGD           & $20.7_{\pm 1.6}$ & $18.8_{\pm 0.9}$ & $86.7_{\pm 1.0}$ & $81.8_{\pm 1.4}$ & $91.8_{\pm 0.7}$ \\
BBB           & $24.5_{\pm 0.7}$ & $17.8_{\pm 0.3}$ & $87.0_{\pm 0.3}$ & $83.4_{\pm 0.4}$ & $91.3_{\pm 0.4}$ \\
Laplace	  & $19.8_{\pm 1.7}$ & $18.8_{\pm 1.0}$ & $86.9_{\pm 1.1}$ & $81.9_{\pm 1.5}$ & $92.0_{\pm 0.8}$ \\
MC-D          & $20.7_{\pm 1.3}$ & ${\bf 17.0}_{\pm 0.6}$ & $88.0_{\pm 0.8}$ & $84.6_{\pm 0.9}$ & $92.1_{\pm 0.7}$ \\
SWAG          & $19.8_{\pm 2.2}$ & ${\bf 16.6}_{\pm 1.0}$ & ${\bf 88.9}_{\pm 1.1}$ & ${\bf 85.3}_{\pm 1.2}$ & $93.0_{\pm 0.9}$ \\
VOGN		  & ${\bf 17.2}_{\pm 1.0}$ & ${\bf 16.5}_{\pm 0.4}$ & ${\bf 89.3}_{\pm 0.6}$ & ${\bf 85.3}_{\pm 0.6}$ & ${\bf 93.5}_{\pm 0.5}$ \\
\rowcolor{gray!10} IVON & ${\bf 17.4}_{\pm 0.8}$ & ${\bf 16.6}_{\pm 0.5}$ & ${\bf 89.2}_{\pm 0.4}$ & ${\bf 85.2}_{\pm 0.6}$ & ${\bf 93.4}_{\pm 0.4}$ \\
              \midrule
              \multicolumn{6}{c}{\textbf{Flowers102}, see~\Cref{fig:flat_d}}\\
              \midrule
              SGD     & $22.1_{\pm 0.5}$ & $20.7_{\pm 0.4}$ & $86.3_{\pm 0.3}$ & $92.1_{\pm 0.2}$ & $75.4_{\pm 0.4}$ \\
BBB           & $22.2_{\pm 0.8}$ & $19.5_{\pm 0.7}$ & $88.2_{\pm 0.7}$ & $93.1_{\pm 0.5}$ & $79.8_{\pm 0.9}$ \\
Laplace	  & $20.5_{\pm 1.1}$ & $20.1_{\pm 0.6}$ & $86.9_{\pm 0.7}$ & $92.4_{\pm 0.4}$ & $76.4_{\pm 1.3}$ \\
MC-D    & $20.3_{\pm 0.8}$ & $19.6_{\pm 1.1}$ & $87.8_{\pm 0.9}$ & $93.0_{\pm 0.7}$ & $78.4_{\pm 1.1}$ \\
SWAG          & $19.5_{\pm 0.8}$ & ${\bf 18.1}_{\pm 0.5}$ & ${\bf 89.3}_{\pm 0.6}$ & ${\bf 93.9}_{\pm 0.4}$ & ${\bf 81.0}_{\pm 0.9}$ \\
VOGN			  & ${\bf 18.1}_{\pm 0.8}$ & ${\bf 18.3}_{\pm 0.3}$ & ${\bf 89.0}_{\pm 0.4}$ & ${\bf 93.8}_{\pm 0.3}$ & ${\bf 80.3}_{\pm 0.7}$ \\
              \rowcolor{gray!10} IVON
                  & ${\bf 17.8}_{\pm 0.5}$ & ${\bf 18.1}_{\pm 0.5}$ & ${\bf 89.0}_{\pm 0.5}$ & ${\bf 93.8}_{\pm 0.3}$ & ${\bf 80.2}_{\pm 0.8}$ \\
                \bottomrule
            \end{tabular}
            }
        \caption{IVON gives good results for OOD detection using a ResNet-20 mode trained on CIFAR-10. The model is evaluated on SVHN and Flowers102. \label{fig:sensitivity-ood}}
        \end{table}

\subsubsection{Out-of-\revision{domain} (OOD) comparisons}

\revision{Next, we consider the OOD case by reusing the CIFAR-10 models on data from a different domain.
While we would expect the model to be certain for correct in-domain predictions, it should be less so for out-of-domain examples.}
This would allow for distinguishing the CIFAR-10 data from OOD samples, for which we use the street view house number (SVHN) \citep{NeWa11} and the 102 Flowers dataset~\citep[Flowers102]{NiZi08}.

\revision{Table~\ref{fig:sensitivity-ood} shows that IVON is
  consistently better at distinguishing OOD examples from SVHN and
  Flowers102 from in-domain CIFAR-10 data.
These results are further illustrated by the predictive entropy plots in
\Cref{fig:flat_c,fig:flat_d}. In these plots, for in-domain data (shown in red), IVON's histogram has a similarly high peak as SGD, but for out-of-domain data (shown in gray) it is much more spread out than SGD.
While the other Bayesian deep learning method's
histograms are also spread out for OOD data, they struggle to achieve a high
peak for in-domain data. Overall, IVON's histogram has the most clear  
separation between in-domain data and out-of-domain data.}
As illustrated in \Cref{fig:in-between}, \revision{IVON's predictive
posterior also gives good in-between uncertainty which has been challenging
for other variational methods~\citep{foong2019between}}.
We show further distribution shift experiments
in~\Cref{app:distshift}.

\label{subsubsec:exp-multi-mcsamples}
\begin{figure}[t!]
	\centering
	\subfigure{
		\includegraphics[width=.47\linewidth]{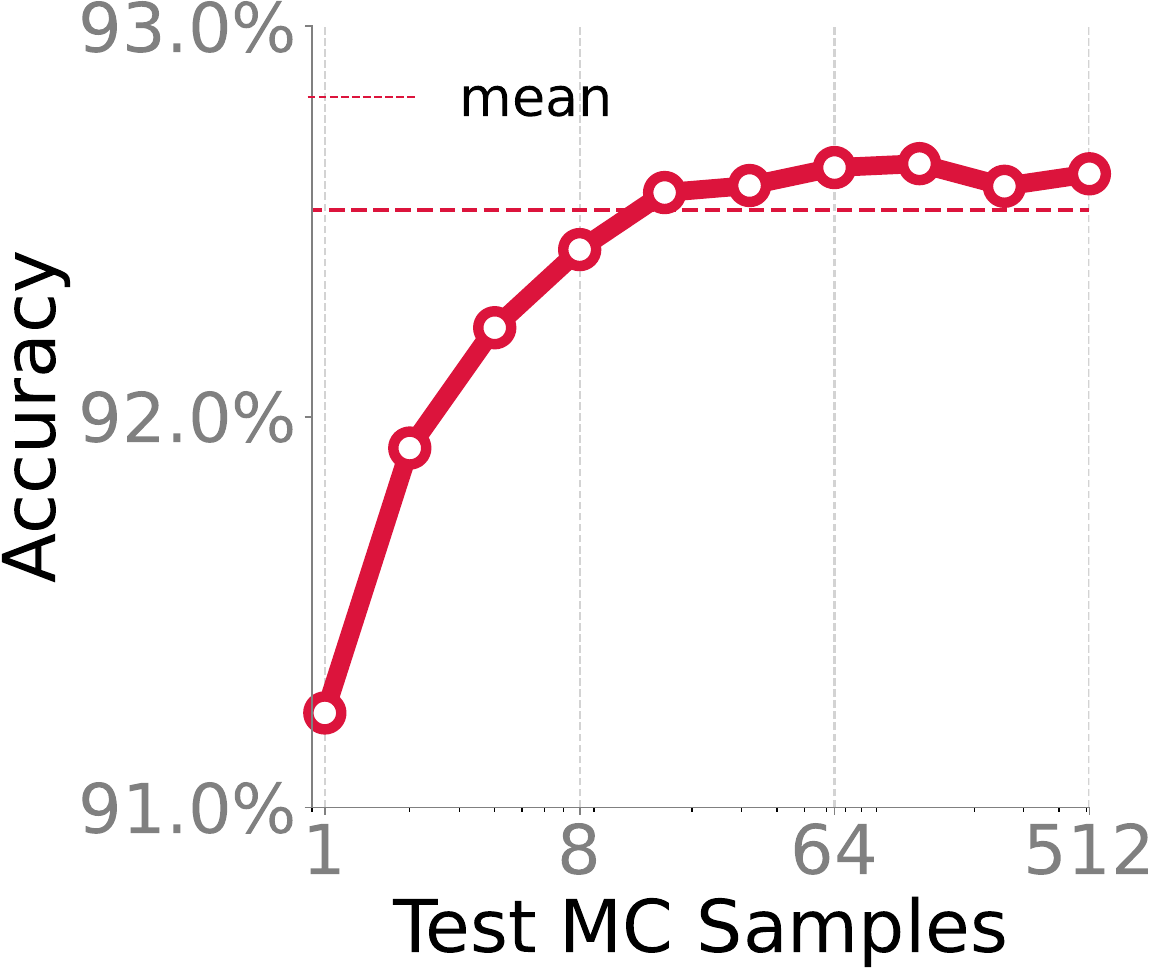}
	}
	\subfigure{
		\includegraphics[width=.47\linewidth]{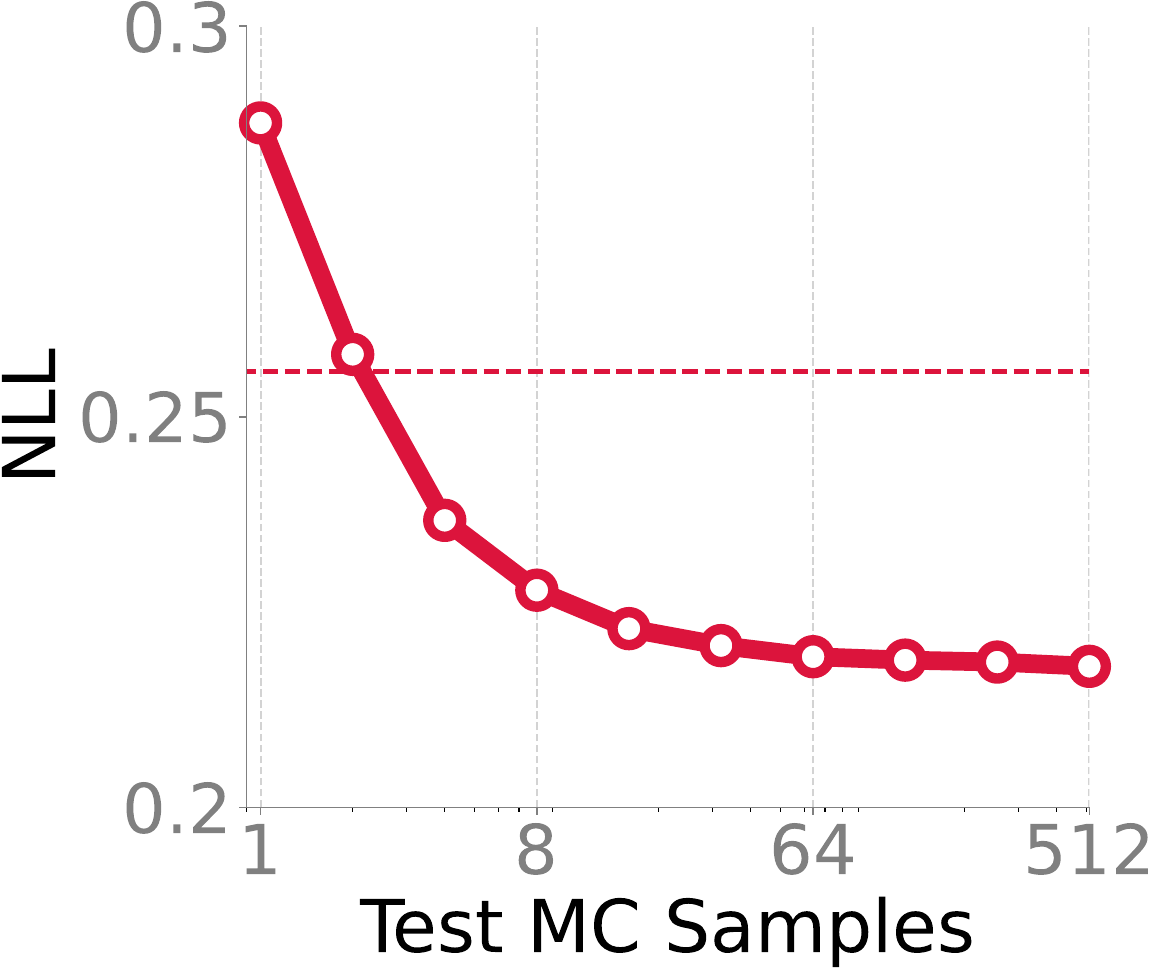}
	}
	\subfigure{
        \subfigure{
            \includegraphics[width=.47\linewidth]{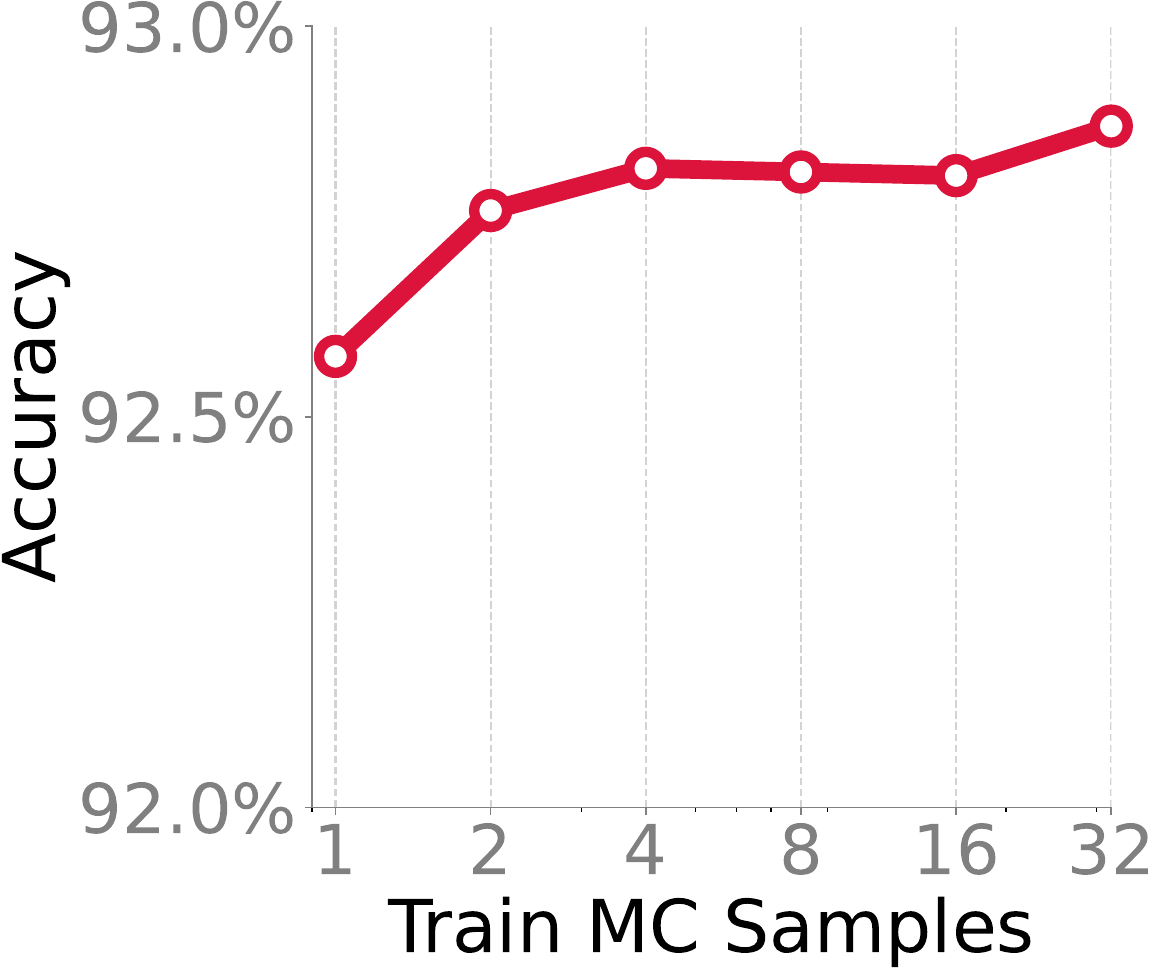}
        }
        \subfigure{
            \includegraphics[width=.47\linewidth]{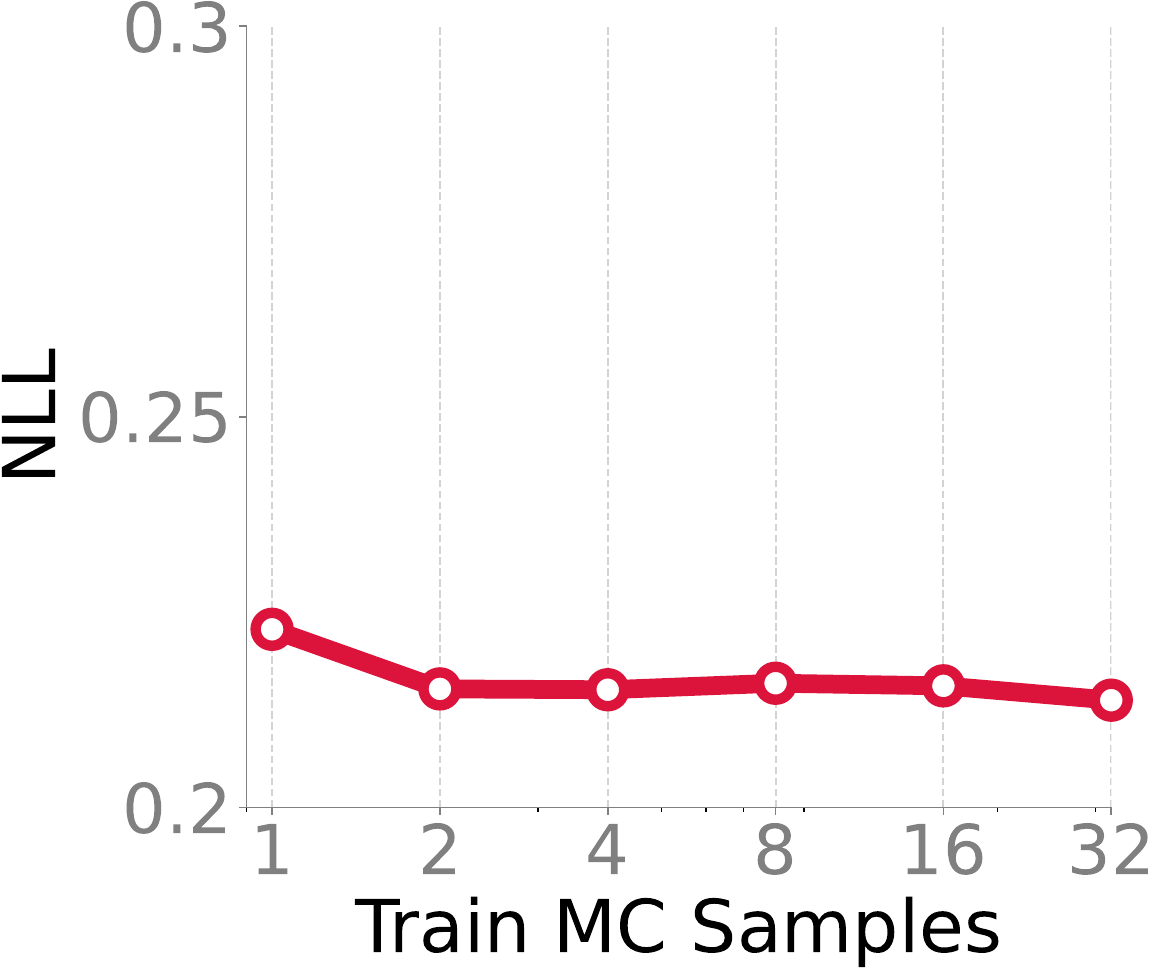}
        }
	}
	\caption{\revision{Using more MC samples during inference (top
            row) or training (bottom row) can
            improve both accuracy and NLL, here plotted for ResNet-20 on CIFAR-10.\label{fig:mcsamples}}}
\end{figure}
\subsubsection{MC samples for averaging}

We find consistent improvements when using more MC samples both during training and inference, but eventually improvements saturate and deliver diminishing returns.
\Cref{fig:mcsamples} (top row) shows that, for ResNet-20 on CIFAR-10, multiple test samples for prediction improves over the prediction using only the mean, especially in terms of NLL.
Similarly, using multiple samples during training improves both accuracy and uncertainty (bottom row in~\Cref{fig:mcsamples}).

\subsubsection{NeurIPS 2021 Competition}
\revision{An earlier version of IVON won first place in both tracks of the
  NeurIPS 2021 competition on approximate inference in Bayesian deep
  learning~\citep{WiIz22}. The methods were evaluated by their 
  difference in predictions to those of a 'ground-truth' posterior
  computed on hundreds of TPUs.
  The winning solution used
  Multi-IVON (same as~\Cref{tab:bdl-indomain}). 
  We summarize the results of the challenge and the
  differences of the earlier version to~\Cref{alg:ivon} in~\Cref{app:challenge}.}

\subsection{Finetuning and Model Merging} 
\label{subsec:benefit-model-merging}

We now demonstrate the usefulness of IVON for finetuning and model-merging. Intuitively, the learned variance $\vsigma^2$ (used to adapt the learning rate) should improve finetuning by favoring directions that are flatter.
It should also be useful to improve Fisher/Hessian-based model merging, for instance, those discussed by \citet{DaMo23}.

\subsubsection{Finetuning pretrained language models}
\label{subsec:benefit-finetune-mlm}
\begin{table}[t!]\centering
    \setlength{\tabcolsep}{3pt}
\resizebox{\linewidth}{!}{\begin{tabular}{lcccccccc}
    \toprule
     & \small MNLI  & \small QNLI &\small QQP &\small RTE &\small SST2 &\small MRPC & \small CoLA &\small STS-B \\
     \toprule
     \multicolumn{9}{c}{DeBERTAv3 (440M params)} \\
    \midrule
    AdamW & 91.3 & {\bf 95.7} & {\bf 93.1} & 91.0 & 96.5 & 91.0 & 74.8 & 92.4\\
    \rowcolor{gray!10} IVON@mean & {\bf 91.6} & {\bf 95.7} & 93.0 & {\bf 91.7} & {\bf 96.9} & {\bf 91.9} & {\bf 75.1} & {\bf 92.6}\\ 
    \hline
    AdamW$^\dagger$ & 91.8 & 96.0 & 93.0 & 92.7 & 96.9 & 92.2 & 75.3 & 93.0\\ \midrule
     \multicolumn{9}{c}{RoBERTa (125M params)} \\ \midrule
		AdamW & 87.7 & {\bf 92.8 } & {\bf 90.9} & {\bf 80.9} & 94.8 & 85.8 & {\bf 63.6 } &90.6 \\
		\rowcolor{gray!10} IVON@mean & {\bf 87.8} & 92.6 & 90.8 & 80.6 & {\bf 95.0} & {\bf 87.3} & 63.3 & {\bf 90.8} \\
    \bottomrule
\end{tabular}}
   \caption{(Top) IVON generally gives better results compared to AdamW for finetuning on DeBERTAv3$_\text{large}$. Better numbers are reported by~\citet{HeGa23} (indicated by AdamW$^\dagger$) but we are not able to reproduce them. (Bottom) Similar results for RoBERTa$_\text{base}$ where performances are comparable but IVON is marginally better on average.}
\label{tab:glue_avg}
\end{table}

\revision{We compare performance on finetuning of a large masked-language model DeBERTAv3~\citep{HeGa23} using AdamW and IVON on GLUE~\citep{WaSi18}. Similarly to previous work~\citep{DeCha18}, we do not include the WNLI dataset.}
DeBERTAv3 has 440M parameters and we finetune the full model using a publicly available checkpoint that was initially trained with AdamW.
The results are shown in the top rows of \Cref{tab:glue_avg} where we see that IVON generally improves upon AdamW; all tasks are for classification except STS-B which is a regression tasks. 
For comparison, we also include results from~\citet{HeGa23} (indicated by AdamW$^\dagger$) which show higher scores but we are unable to reproduce them.
Bottom row shows similar results for RoBERTa~\citep{liuOtt19}, where IVON performs more comparably to AdamW but still gives marginally better results on average.
\revision{Experimental details are in~\Cref{app:roberta}.}

\begin{table}[t!]\centering

    \setlength{\tabcolsep}{3pt}
    \small
    
        \begin{tabular}{lrrrrrrr}
    \toprule
    & IMDB & Yelp & RT & SST2 & Amazon & Avg. & Overhead \\
    \midrule
    SG & 93.5 & 97.0 & 89.7 & 92.8 & 96.6 & 93.9 & 100\% \\
    \rowcolor{gray!10} IVON & 93.6 & 96.9 & 89.8 & 92.8 & 96.7 & 94.0 & 0\% \\
    \bottomrule
    \end{tabular}
    \caption{IVON reduces the cost of Hessian-based model-merging while giving comparable results to an existing method which requires an extra post-training run through the full data to compute the square-gradients (SG). For IVON, we simply use $\vh$ obtained during training which has zero overhead (indicated in last column with $0\%$).}
    \label{tab:merging}
    \end{table}

\subsubsection{Merging masked-language models}
\label{sec:merging_results}
\revision{We repeat the experiments by~\citet{DaMo23} to merge finetuned RoBERTa models on} 
IMDB, Amazon, Yelp, RottenTomatoes, and SST2.
Given pretrained weights $\vparam_0$, they use the following method to merge $T$ finetuned models $\vparam_t$,
\begin{equation*}    
   \vparam_{\text{merged}} = \vparam_0 + \smash{\sum_{t=1}^T \frac{\vh_0 + \vh_t}{\vh_0 + \sum_{t'=1}^T \vh_{t'}}(\vparam_t - \vparam_0)},
\end{equation*}
where $\vh_t$ is the diagonal of the Hessian at the $t$'th model. For IVON, we use the vector $\vh$ as the Hessian estimate.
We compare it to the squared gradient (SG) estimator: $\vh_t =
\sum_{i} [\nabla \hat{\ell}_i(\vparam_t)]^2$. The loss $\hat{\ell}_i$
here is the same as $\loss_i$ but uses labels sampled from the
model. The estimator is related to Laplace's method~\citep{DaMo23} and
requires an extra post-training run over the full dataset. For both
methods, we set $\vh_0$ using the SG method because IVON is not used
during pretraining.
We use same settings as~\Cref{app:roberta}.

Table~\ref{tab:merging} shows that both estimates perform similarly, despite IVON not needing an extra pass through the dataset (indicated by $0\%$ overhead in the last column).
We expect the results to get even better when the model is pretrained with IVON, because then $\vh_0$ is also estimated during pretraining.

\subsection{Estimating Generalization and Sensitivity to Data}
\label{subsec:benefit-data-sensitivity}

We demonstrate the usefulness of IVON to faithfully predict generalization performance and also understand sensitivity to training data. The former is useful for training diagnostics and early stopping, while the latter is useful for model understanding and data cleaning. To do so, we simply plug-in the variance estimate $\vsigma^2$ obtained during IVON's training into the influence measures derived by \citet{NiXu23}. We find this simple method to give good results
compared to existing influence function methods~\citep{KoLi17} that are known to be fragile for large deep networks~\citep{basu2021influence} but also are not designed to work during training.

\subsubsection{Predicting generalization Performance}
\label{subsubsec:generalization}
We use IVON to estimate the Leave-One-Out (LOO) Cross-Validation loss as a measure of generalization performance. Given a $\vparam$ during training, \citet{NiXu23} use its prediction error $e_i$ and prediction variance $v_i$ on the $i$'th example to estimate the deviation $\vparam^{\backslash i}$ obtained by removing the same $i$'th example.
This is then used to estimate the LOO loss,
\[
   \text{LOO}(\vtheta) = \sum_{i=1}^N \ell(f_i^{\backslash i}) \approx \sum_{i=1}^N \ell(f_i + v_i e_i),
\]
where $f_i$ and $f_i^{\backslash i}$ are predictions by using $\vparam$ and $\vparam^{\backslash i}$, respectively, for the $i$'th example and we denote $\loss(f_i) = \loss_i(\vparam)$. IVON's variance $\vsigma^2$ is expected to improve the predictive variance $v_i$ (see \Cref{app:sensitivity-method}) and this experiment demonstrates the effectiveness of the variance estimate compared to the Squared-Gradient (SG) estimate used in Adam. 

\begin{figure}[!t]
    \subfigure[IVON]{
        \includegraphics[width=.47\linewidth]{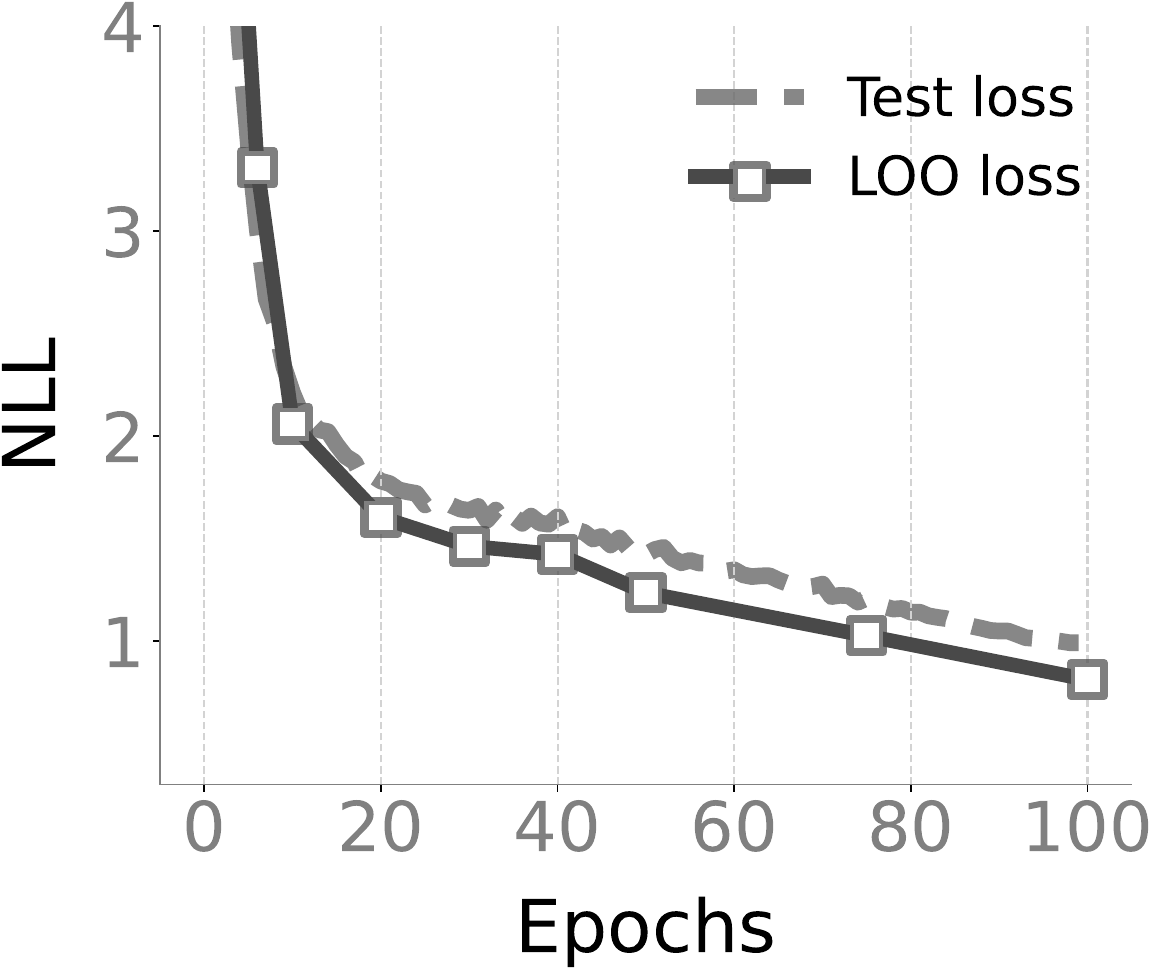}
        }
        \subfigure[AdamW]{
            \includegraphics[width=.47\linewidth]{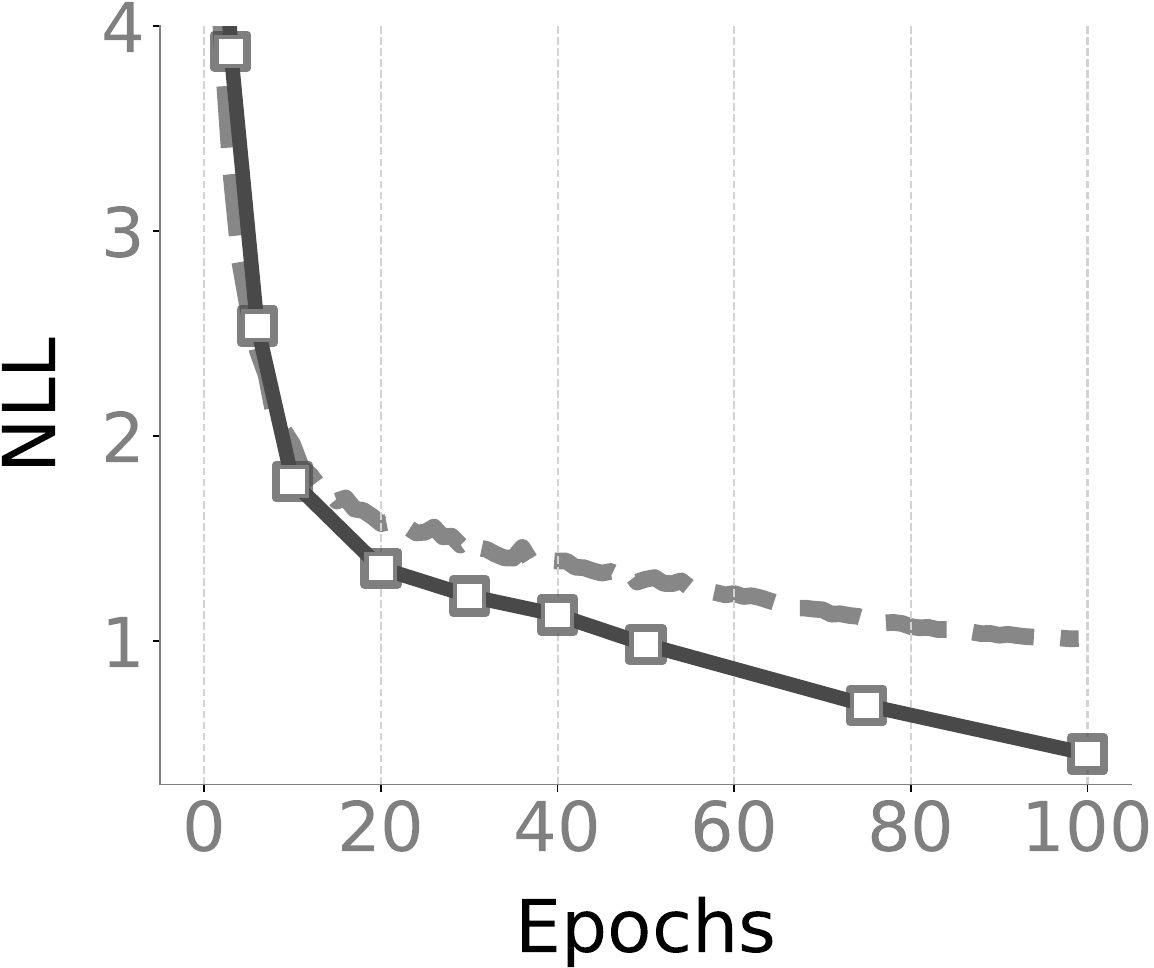}
            }
            \caption{Panel (a) shows that, for ImageNet, IVON's LOO estimate (solid line with square markers) accurately follows the loss trajectory on an unseen test set (dashed line). Panel (b) shows that same for AdamW which is not as good. \label{fig:generalization_prediction}}
          \end{figure}

\Cref{fig:generalization_prediction} shows the results on ImageNet (setup similar to the one shown in \cref{tab:imagenet}). We see that IVON's LOO estimate closely follows the (true) loss on an unseen test set. 
The LOO objective is evaluated using sensitivities calculated from IVON during training. 
Similar estimates obtained with AdamW do not work as well, for instance, they converge close to 0 which reflects the training loss more than the test loss. Details of the method used for AdamW are given in \Cref{app:sensitivity-method}.
Each plot shows LOO at regular intervals indicated with the markers. 
Additional results for many other architectures on CIFAR-10 are given in~\Cref{app:exp-predict-gen}.

\begin{figure*}[t] 
	\centering
        \includegraphics[width=0.95\linewidth]{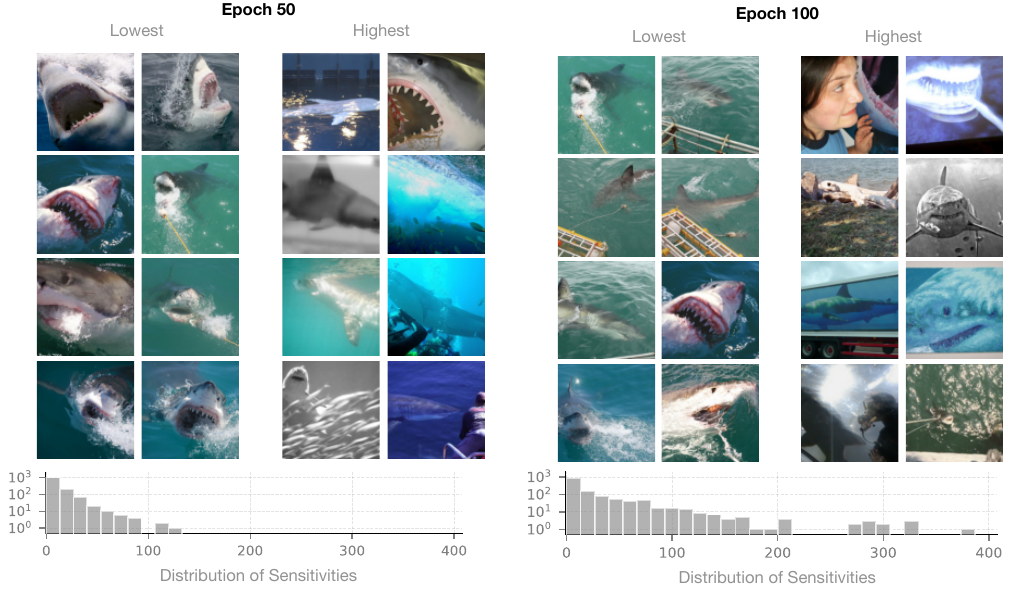}
    \caption{We use IVON to estimate sensitivity of ResNet-50 to examples in the ImageNet dataset, and visualize high and low sensitivity images for the ``great white shark'' class at two epochs of 50 and 100 respectively. Low-sensitivity examples are mostly typical shark images, while high-sensitivity ones are unusual images, possibly containing mislabeled or ambiguous examples. For instance, one picture shows more prominently the face of a woman than the shark. The high-sensitivity
    examples also continue to evolve when going from 50 to 100 epochs (the distribution of sensitivities flattens), perhaps indicating that the model tends to learn them a bit later in the training. \label{fig:sensi}}
\end{figure*}

\subsubsection{Sensitivity to Training Data}
\label{subsubsec:dataanalysis}
We present qualitative results to estimate sensitivity of the model to its training data. Similarly to \citet{NiXu23}, we define the sensitivity measure to be the absolute value of $v_ie_i$. Higher values suggest higher sensitivities.
Because the sensitivity estimates are obtained by multiplying $e_i$ and $v_i$, we expect low sensitivity images to be those predicted correctly with high confidence, that is, they should be \emph{typical} images. In contrast, high sensitivity images are the ones with high prediction error and/or high variance, which in general would be \emph{atypical} ones (mislabeled or ambiguous).

\Cref{fig:sensi} illustrates low- and high-sensitivity images for the ``great white shark'' class on ImageNet (details in \Cref{app:sensitivity-method}). We show two epochs (50 and 100, respectively). Already at 50 epochs, low-sensitivity examples show the regular shark pattern, but high-sensitivity examples keep appearing as the training progresses (the distribution of sensitivities flattens). At epoch 100, we see some clear examples containing atypical images, for instance, a picture of a woman's face featured more prominently than the shark.

\section{Discussion and Limitations}
We show the effectiveness of variational learning for training large networks. 
Especially our results on GPT-2 and other LLMs are first of their kind and clearly demonstrate the potential that variational learning holds. 
We also discussed many new use cases where we consistently find benefits by switching to a variational approach. 
We expect our results to be useful for future work on showing \revision{the} effectiveness of Bayesian learning in general. 

Although we borrow practical tricks from deep learning, not all of
them are equally useful for IVON, for example, we find that IVON does
not go well with batch normalization layers~\citep{IoSz15}. Future
research should explore this limitation and investigate the reasons
behind \revision{the} effectiveness of some practical tricks. Using MC samples in
variational learning increases the computation cost and we believe it
is difficult to fix this problem. For this, deterministic versions of the
variational objective can be useful, for example, those
discussed by~\citet{MoKh23} but this is a potential direction for future research.

IVON can be easily modified to learn flexible posterior
forms~\citep{LiKh19b}. Our Multi-IVON method in this paper uses a
simple mixture distribution, but we expect further improvements by
using other types of mixtures and also by learning the mixing
distribution. We expect this aspect of IVON to help researchers further investigate the benefits of Bayesian principles to improve deep learning.

\section*{Acknowledgements}
This work is supported by JST CREST Grant Number JPMJCR2112.
This research work has been funded by the German Federal Ministry of Education and Research and the Hessian Ministry of Higher Education, Research, Science and the Arts within their joint support of the National Research Center for Applied Cybersecurity ATHENE.
Y.~Shen and D.~Cremers are supported by the Munich Center for Machine Learning (MCML) and the ERC
Advanced Grant SIMULACRON.
Some of the experiments were carried out with the TSUBAME3.0
supercomputer at the Tokyo Institute of Technology.

We thank Happy Buzaaba for first experiments on finetuning transformers with IVON and 
Keigo Nishida, Falko Helm and Hovhannes Tamoyan for feedback on a draft of this paper.
Finally, we thank the organizers of the NeurIPS 2021 Competition on
Approximate Inference in Bayesian Deep Learning.

\section*{Impact Statement}
This paper presents work whose goal is to advance the field
of Machine Learning. There are many potential societal consequences of our work, 
none which we feel must be specifically highlighted here.

\bibliography{references}
\bibliographystyle{icml2024}

\newpage
\appendix
\onecolumn

\section{Practical Guideline for Choosing IVON Hyperparameters}
\label{subsec:guide}
 
To facilitate the usage of IVON, we provide here some practical guidelines for choosing hyperparameters and refer to their notations from Algorithm~\ref{alg:ivon}.

\paragraph{Learning rate schedule $\alpha_t$.} For ResNets, the initial learning rate of IVON can be set to the same value that works well for SGD, or slightly larger.
For Transformers, we have found larger learning rates to work well, such as 0.1 for finetuning RoBERTa~\citep{liuOtt19}, or 0.2 for pretraining GPT-2~\citep{RaWu2019} with 355M parameters.
\revision{Typical learning rate schedules like linear decay or cosine annealing work well for IVON.
We have found decaying the learning rate to $0$ to work best for pretraining GPT-2, better than decaying it to the initial learning rate divided by 10 as suggested by~\citet{HoBo22}.}

\paragraph{Effective sample size $\lambda$.} Setting this to the size of training dataset ($\lambda = N$) in Eq.~\eqref{eq:ivonobjective} is a good starting point. 
This recovers the standard evidence lower bound objective for variational learning.
\revision{Setting it smaller is equivalent to increased temperature and setting it higher to decreased temperature.}
In our experiments we mostly set $\lambda \approx N$, except for finetuning transformers on very small datasets 
where we notice larger $\lambda$ can improve performance and stabilize the short training. \revision{As seen from line~8 in~\Cref{alg:ivon},
the choice of $\lambda$ directly influences the posterior variance and too small values may lead to a high variance and unstable training
whereas too large values may lead to a collapsed posterior that offers little benefits.}

\paragraph{Weight decay $\delta$.} For ResNets, the weight decay of IVON can be set to the same values that work well for SGD or Adam. 
For Transformers, we have found smaller values, such as $10^{-5}$, which we use for finetuning, or $10^{-6}$, which we use for pretraining, to work well for weight decay. 
Larger values are feasible when using a quadratic penalty biased to the initialization of the model for finetuning.

\paragraph{Gradient momentum $\beta_1$.} Setting $\beta_1 = 0.9$ tends to work well, similar to SGD or Adam. 
This plays a similar role as the gradient momentum in other optimizers so we expect the good settings to be similar.

\paragraph{Hessian momentum $\beta_2$.} The Hessian momentum needs to be set rather close to one, \revision{for example,} $\beta_2 = 1 - 10^{-5}$ worked well in our experiments. 
The Hessian momentum \revision{in theory} is given by $\beta_2 = 1 - \lambda^{-1} N \rho$, where $\rho$ is the step-size of natural gradient descent. 
If $\beta_2$ is set too small, for example, $0.999$ or $0.9999$ the
\revision{training} can sometimes become unstable. 

\paragraph{Hessian initialization $h_0$.} Along with the effective sample size $\lambda$, the Hessian initialization $h_0$ controls the noise at initialization. 
Typically values between $0.01$ and $1$ work well in practice but also smaller values like $0.001$ have shown good results. 
Large values of $h_0$ correspond to more concentrated and deterministic initial posterior and can help stabilizing the training, but this can lead to poorer results. 
It can be helpful to monitor the statistics of the Hessian vector $\vh$ during training, to see whether a reasonable covariance is being learned.

\paragraph{Batch size, training epochs.} Typical batch sizes and training epochs that work well for SGD \revision{and AdamW} tend to also work well for IVON. 
\revision{For example, our GPT-2 results in~\Cref{fig:teaser_a} use the same batch size and number of epochs for IVON and AdamW.}
This said, we observe that longer training and larger batch size seems to benefit IVON more than SGD, possibly because this would further improve the Hessian estimate.

\paragraph{Clip radius $\xi$.} When training transformers,
element-wise gradient clipping can stabilize the training. 
A clip-radius of $\xi=10^{-3}$ worked well in practice. When picking a smaller clip-radius, one often requires a larger learning rate.

\section{Ablation Studies}
\label{app:exp-ablation}

\subsection{Computational Efficiency of IVON}
\label{app:efficiency}
\begin{table}[ht!]\centering
    \setlength{\tabcolsep}{3pt}
    
      \begin{tabular}{lcccc@{\hskip .25in}cccc}
    \toprule
        & \multicolumn{4}{c}{Runtime (hrs)} & \multicolumn{4}{c}{Memory (GB)} \\
        & AdamW & SGD & VOGN & IVON & AdamW & SGD & VOGN & IVON \\ \midrule
        ResNet-20 & 0.38 & 0.38 & 0.68 & 0.38 & 1.7 & 1.7 & 2.0 & 1.7 \\
        GPT-2 (125M) & 15.0 & - & - & 18.5 & 21.8 & - & - & 23.2 \\
        GPT-2 (355M) & 37.5 & - & - & 44.7 & 23.7 & - & - & 27.7 \\ \bottomrule
      \end{tabular}
    \caption{Runtime and memory for CIFAR-10 classification results
      with ResNet-20 and pretraining GPT-2 on OpenWebText. IVON has a small overhead for larger models which might be due to 
      the additional weight sampling and a not fully optimized implementation.}
    \label{tab:runtime}
    \end{table}
\begin{table}[h!]
\centering

\begin{tabular}{lcccccc}
\toprule
 & Acc. $\uparrow$ & NLL $\downarrow$ & ECE $\downarrow$  & Brier $\downarrow$ & Mem $\downarrow$ \\
\midrule
SG & $88.81_{\pm 0.31}$ & $0.464_{\pm 0.020}$ & $0.070_{\pm 0.004}$ & $0.180_{\pm 0.006}$ & {\bf 363MB} \\
GGN & $92.37_{\pm 0.23}$ & $0.226_{\pm 0.005}$ & ${\bf 0.008}_{\pm 0.001}$ & $0.111_{\pm 0.003}$ & 645MB \\
\rowcolor{gray!10} Reparam.    & ${\bf 92.64}_{\pm 0.13}$ & ${\bf 0.219}_{\pm 0.005}$ & ${\bf 0.009}_{\pm 0.002}$ & ${\bf 0.107}_{\pm 0.002}$ & {\bf 363MB} \\
\bottomrule
\end{tabular}
\caption{IVON's reparameterization-trick-based Hessian estimator has better accuracy and uncertainty than other Hessian estimators at low computational cost, here for ResNet-20 on CIFAR-10.}
\label{tab:ablation-ivon-hess}
\end{table}

 The computational budget required by IVON is similar to standard deep learning optimizers.
To validate its efficiency empirically, we measure the run time and peak GPU memory usage 
for image classification experiments on CIFAR-10 with ResNet-20 \citep{HeZh16} with an identical setup except 
for the choice of optimizer. 
\Cref{tab:runtime} shows that IVON has similar computational costs as SGD and AdamW.
However, we find a slight overhead when training larger models like GPT-2 as shown in~\Cref{fig:teaser_a}, potentially because of the additional sampling step 
\revision{and unoptimized implementation}.

\subsection{Comparison of Hessian estimators}
\label{app:exp-ablation-hess}
IVON's efficiency is enabled by estimating $\vddl$ with the \revision{reparameterization-trick-based}
estimator in~\Cref{eq:lh_estimator}.
Here, we compare this estimator to the two squared-gradient estimators
discussed in the previous section: 
1) the Squared Gradient (SG) estimator which uses the square of mini-batch gradients
$\smash{\widehat \vh \leftarrow \widehat \vg^2}$ used in Vprop and Vadam~\citep{KhNi18};
2) the Gauss-Newton (GN) estimator which uses per-sample squared gradients,
$\smash{\widehat \vh \leftarrow \frac{1}{|\mathrm{B}|}\sum_{i \in \mathrm{B}} \big[\nabla \ell_i(\vparam + \vsigma \vepsilon) \big]^2}$ used in VOGN~\citep{OsSw19}.
One drawback of the GN estimator is that per-example gradients require
significant overhead, since the backpropagation process of typical
deep learning frameworks only computes  an averaged mini-batch gradient $\widehat \vg$.

Table~\ref{tab:ablation-ivon-hess} shows results for training ResNet-20 on CIFAR-10 with these estimators. 
We observe that the \revision{reparameterization} estimator provides best performance. 
The squared gradient estimator is similarly efficient but underperforms, whereas Gauss-Newton incurs significant overhead in GPU memory and time usage 
without large benefits in test performance. 

\subsection{Hessian initialization}
\label{app:hess-init-ablation}
We perform an ablation over the Hessian initialization $h_0$ used in \Cref{alg:ivon}. The model is a ResNet-20 trained on CIFAR-10. The results are summarized in the following \Cref{tab:hessabl}.
\begin{table}[htbp]
  \centering
  \begin{tabular}{lccccc}
  \toprule
  $h_0$ & Acc. $\uparrow$ & NLL $\downarrow$ & ECE $\downarrow$ & Brier $\downarrow$ \\ \midrule
  0.001 & $9.89_{\pm 00.24}$ & $2.3027_{\pm 0.0001}$ & $0.0033_{\pm 0.0024}$ & $0.9000_{\pm 0.0000}$ \\
  0.002 & $43.95_{\pm 31.24}$ & $1.4919_{\pm 0.7403}$ & $0.0495_{\pm 0.0410}$ & $0.6328_{\pm 0.2440}$ \\
  0.005 & $82.78_{\pm 00.50}$ & $0.5612_{\pm 0.0093}$ & $0.1098_{\pm 0.0024}$ & $0.2660_{\pm 0.0049}$ \\
  0.01 & $86.74_{\pm 00.15}$ & $0.4477_{\pm 0.0056}$ & $0.1019_{\pm 0.0006}$ & $0.2108_{\pm 0.0028}$ \\
  0.02 & $89.60_{\pm 00.35}$ & $0.3598_{\pm 0.0082}$ & $0.0871_{\pm 0.0032}$ & $0.1690_{\pm 0.0034}$ \\
  0.05 & $91.72_{\pm 00.09}$ & $0.2777_{\pm 0.0035}$ & $0.0604_{\pm 0.0030}$ & $0.1318_{\pm 0.0015}$ \\
  0.1 & $92.25_{\pm 00.17}$ & $0.2419_{\pm 0.0026}$ & $0.0388_{\pm 0.0031}$ & $0.1174_{\pm 0.0017}$ \\
  0.2 & $92.44_{\pm 00.18}$ & $0.2243_{\pm 0.0043}$ & $0.0186_{\pm 0.0012}$ & $0.1106_{\pm 0.0020}$ \\
  0.5 & ${\bf 92.71}_{\pm 00.07}$ & ${\bf 0.2173}_{\pm 0.0045}$ & ${\bf 0.0066}_{\pm 0.0015}$ & ${\bf 0.1069}_{\pm 0.0012}$ \\
  1 & $92.61_{\pm 00.08}$ & $0.2329_{\pm 0.0042}$ & $0.0154_{\pm 0.0017}$ & $0.1111_{\pm 0.0011}$ \\
  2 & $92.46_{\pm 00.23}$ & $0.2411_{\pm 0.0074}$ & $0.0251_{\pm 0.0032}$ & $0.1125_{\pm 0.0025}$ \\
  5 & $92.34_{\pm 00.29}$ & $0.2613_{\pm 0.0151}$ & $0.0331_{\pm 0.0036}$ & $0.1167_{\pm 0.0050}$ \\
  \bottomrule
  \end{tabular}
  \caption{The initialization of the Hessian $\vh$ in \Cref{alg:ivon} can be important to achieve the best results.}
  \label{tab:hessabl}
  \end{table}

\section{Experimental Details}
\subsection{Pretraining GPT-2 Models on OpenWebText}
\label{app:gpt2_details}
We pretrain GPT-2 models~\citep{RaWu2019} on OpenWebText~\citep{GoCo19} for multiple epochs and around 49.2B tokens in total using a batch size of 480 which is achieved by gradient accumulation.
We train on 8 \revision{NVIDIA} A100 GPUs \revision{with 40GB GPU memory each} for up to \revision{three} days.
We use 2,000 warmup steps, 100,000 training steps in total, and evaluate every 1,000 steps on a held-out set.
\revision{Each validation step is shown in~\Cref{fig:teaser_a}.}
The learning rate is decayed to $0$, which we have found to work better than $1/10$-times the initial learning rate for both AdamW and IVON.
This is recommended in prior work~\citep{HoBo22}.
For IVON, we use an initial learning rate of $0.3$ for the 125M parameter checkpoint, $0.2$ for the 355M parameter checkpoint, and $0.15$ for the 773M parameter checkpoint.
\revision{Note, that we do not rescale by $\vh_0$ and $\delta$ in this case, because element-wise clipping is used.}
We use $\beta_1=0.9$, $\beta_2 = 1-10^{-5}$, $h_0 = 0.001$ and a weight decay factor of $10^{-6}$, as well as element-wise clipping of $10^{-3}$.
These hyperparameters were found by grid search on a smaller model and it is possible that better hyperparameter configurations exist.
\revision{We train with a single MC sample}.

For training GPT-2 with AdamW, we use an initial learning rate of $6\cdot 10^{-4}$, $\beta_1=0.9$, $\beta_2=0.95$ and a weight decay of $0.1$.
This follows the hyperparameters used in prior works~\citep{LiLi23}.
We follow the implementation in \url{https://github.com/karpathy/nanoGPT/}, which uses GeLU activations~\citep{HeGi16} and does not use dropout~\citep{SriHi14} and biases during pretraining.

\begin{table}[t!]\centering
\setlength{\tabcolsep}{3pt}

\begin{tabular}{llccccc}
\toprule
 &  & Acc. $\uparrow$ & Top-5 Acc. $\uparrow$ & NLL $\downarrow$ & ECE $\downarrow$ & Brier $\downarrow$ \\
\midrule
                                                                                                                        & AdamW & $90.04_{\pm 0.27}$ & $99.62_{\pm 0.03}$ & $0.589_{\pm 0.018}$ & $0.074_{\pm 0.002}$ & $0.170_{\pm 0.004}$ \\
                                                                                                                        & AdaHessian & $91.46_{\pm 0.06}$ & $99.71_{\pm 0.02}$ & $0.477_{\pm 0.018}$ & $0.061_{\pm 0.001}$ & $0.144_{\pm 0.001}$ \\
                                                                                                                        & SGD & $91.86_{\pm 0.14}$ & ${\bf 99.70}_{\pm 0.08}$ & $0.288_{\pm 0.015}$ & $0.040_{\pm 0.004}$ & $0.126_{\pm 0.003}$ \\
                                                                                                                        \rowcolor{gray!10} \cellcolor{white}& IVON@mean & $92.53_{\pm 0.04}$ & ${\bf 99.77}_{\pm 0.05}$ & $0.256_{\pm 0.005}$ & $0.034_{\pm 0.001}$ & $0.115_{\pm 0.001}$ \\
                                                                                                                        \rowcolor{gray!10} \cellcolor{white}\multirow{-5}*{\begin{tabular}{l} \cellcolor{white} {ResNet-20}\\ \cellcolor{white} (272k params) \end{tabular}} & IVON & ${\bf 92.71}_{\pm 0.07}$ & ${\bf 99.78}_{\pm 0.03}$ & ${\bf 0.219}_{\pm 0.002}$ & ${\bf 0.008}_{\pm 0.001}$ & ${\bf 0.108}_{\pm 0.001}$ \\
\cmidrule{1-7}
                                                                                                                         & AdamW & $92.41_{\pm 0.26}$ & ${\bf 99.72}_{\pm 0.04}$ & $0.594_{\pm 0.022}$ & $0.062_{\pm 0.002}$ & $0.135_{\pm 0.005}$ \\
                                                                                                                         & AdaHessian & ${\bf 92.95}_{\pm 0.87}$ & ${\bf 99.72}_{\pm 0.14}$ & $0.514_{\pm 0.028}$ & $0.056_{\pm 0.006}$ & $0.124_{\pm 0.014}$ \\
                                                                                                                         & SGD & $92.54_{\pm 0.30}$ & $99.62_{\pm 0.04}$ & $0.328_{\pm 0.008}$ & $0.050_{\pm 0.003}$ & $0.123_{\pm 0.003}$ \\
                                                                                                                         \rowcolor{gray!10} \cellcolor{white}& IVON@mean & ${\bf 93.31}_{\pm 0.31}$ & ${\bf 99.74}_{\pm 0.03}$ & $0.282_{\pm 0.014}$ & $0.042_{\pm 0.003}$ & $0.110_{\pm 0.004}$ \\
                                                                                                                         \rowcolor{gray!10} \cellcolor{white}\multirow{-5}*{\begin{tabular}{l} \cellcolor{white} {DenseNet-121}\\ \cellcolor{white} (1M params) \end{tabular}} & IVON & ${\bf 93.53}_{\pm 0.26}$ & ${\bf 99.78}_{\pm 0.04}$ & ${\bf 0.200}_{\pm 0.007}$ & ${\bf 0.009}_{\pm 0.001}$ & ${\bf 0.096}_{\pm 0.003}$ \\
\cmidrule{1-7}
                                                                                                                                & AdamW & $92.39_{\pm 0.27}$ & $99.69_{\pm 0.05}$ & $0.653_{\pm 0.024}$ & $0.064_{\pm 0.003}$ & $0.137_{\pm 0.005}$ \\
                                                                                                                                & AdaHessian & ${\bf 93.76}_{\pm 0.25}$ & ${\bf 99.78}_{\pm 0.03}$ & $0.431_{\pm 0.021}$ & $0.049_{\pm 0.002}$ & $0.109_{\pm 0.004}$ \\
                                                                                                                                & SGD & $93.70_{\pm 0.15}$ & $99.66_{\pm 0.08}$ & $0.298_{\pm 0.010}$ & $0.045_{\pm 0.001}$ & $0.107_{\pm 0.002}$ \\
                                                                                                                                \rowcolor{gray!10} \cellcolor{white}& IVON@mean & ${\bf 93.99}_{\pm 0.08}$ & ${\bf 99.80}_{\pm 0.03}$ & $0.259_{\pm 0.008}$ & $0.042_{\pm 0.001}$ & $0.100_{\pm 0.002}$ \\
                                                                                                                                \rowcolor{gray!10} \cellcolor{white}\multirow{-5}*{\begin{tabular}{l} \cellcolor{white} {PreResNet-110}\\ \cellcolor{white} (deep, 4M params) \end{tabular}} & IVON & ${\bf 94.02}_{\pm 0.14}$ & ${\bf 99.84}_{\pm 0.03}$ & ${\bf 0.180}_{\pm 0.003}$ & ${\bf 0.010}_{\pm 0.001}$ & ${\bf 0.087}_{\pm 0.001}$ \\
\cmidrule{1-7}
                                                                                                                             & AdamW & $92.40_{\pm 0.32}$ & $99.69_{\pm 0.05}$ & $0.676_{\pm 0.006}$ & $0.064_{\pm 0.003}$ & $0.137_{\pm 0.005}$ \\
                                                                                                                             & AdaHessian & $88.66_{\pm 1.51}$ & $99.38_{\pm 0.13}$ & $0.569_{\pm 0.037}$ & $0.081_{\pm 0.008}$ & $0.190_{\pm 0.023}$ \\
                                                                                                                             & SGD & $94.03_{\pm 0.14}$ & $99.72_{\pm 0.03}$ & $0.282_{\pm 0.009}$ & $0.043_{\pm 0.002}$ & $0.101_{\pm 0.003}$ \\
                                                                                                                             \rowcolor{gray!10} \cellcolor{white}& IVON@mean & ${\bf 94.17}_{\pm 0.08}$ & ${\bf 99.78}_{\pm 0.04}$ & $0.305_{\pm 0.007}$ & $0.045_{\pm 0.001}$ & $0.102_{\pm 0.002}$ \\
                                                                                                                             \rowcolor{gray!10} \cellcolor{white}\multirow{-5}*{\begin{tabular}{l} \cellcolor{white} {ResNet-18}\\ \cellcolor{white} (wide, 11M params) \end{tabular}} & IVON & ${\bf 94.32}_{\pm 0.13}$ & ${\bf 99.84}_{\pm 0.03}$ & ${\bf 0.175}_{\pm 0.002}$ & ${\bf 0.010}_{\pm 0.001}$ & ${\bf 0.084}_{\pm 0.001}$ \\
\bottomrule
\end{tabular}
\caption{IVON results on CIFAR-10 compared with various baseline optimizers using convolutional networks with different widths and depths. IVON@mean denotes point estimate results evaluated at the mean of IVON posterior.}
\label{tab:resnets:cifar10}
\end{table}

\begin{table}[b!]\centering
\setlength{\tabcolsep}{3pt}

\begin{tabular}{llccccc}
\toprule
 &  & Acc. $\uparrow$ & Top-5 Acc. $\uparrow$ & NLL $\downarrow$ & ECE $\downarrow$ & Brier $\downarrow$ \\
\midrule
                                                                                                                        & AdamW & $60.76_{\pm 0.47}$ & $86.81_{\pm 0.48}$ & $1.931_{\pm 0.044}$ & $0.202_{\pm 0.004}$ & $0.580_{\pm 0.008}$ \\
                                                                                                                        & AdaHessian & $64.19_{\pm 0.28}$ & $88.68_{\pm 0.39}$ & $1.612_{\pm 0.033}$ & $0.167_{\pm 0.007}$ & $0.521_{\pm 0.004}$ \\
                                                                                                                        & SGD & $67.23_{\pm 0.35}$ & $90.75_{\pm 0.11}$ & $1.173_{\pm 0.021}$ & $0.059_{\pm 0.008}$ & $0.441_{\pm 0.005}$ \\
                                                                                                                        \rowcolor{gray!10} \cellcolor{white}& IVON@mean & ${\bf 67.87}_{\pm 0.55}$ & $90.95_{\pm 0.10}$ & $1.168_{\pm 0.012}$ & $0.069_{\pm 0.007}$ & $0.438_{\pm 0.005}$ \\
                                                                                                                        \rowcolor{gray!10} \cellcolor{white}\multirow{-5}*{\begin{tabular}{l} \cellcolor{white} {ResNet-20}\\ \cellcolor{white} (272k params) \end{tabular}} & IVON & ${\bf 68.28}_{\pm 0.50}$ & ${\bf 91.27}_{\pm 0.05}$ & ${\bf 1.113}_{\pm 0.010}$ & ${\bf 0.018}_{\pm 0.003}$ & ${\bf 0.425}_{\pm 0.005}$ \\
\cmidrule{1-7}
                                                                                                                         & AdamW & $65.47_{\pm 0.93}$ & $88.74_{\pm 0.80}$ & $2.967_{\pm 0.104}$ & $0.264_{\pm 0.007}$ & $0.587_{\pm 0.015}$ \\
                                                                                                                         & AdaHessian & $71.02_{\pm 0.57}$ & $92.00_{\pm 0.17}$ & $2.379_{\pm 0.038}$ & $0.222_{\pm 0.005}$ & $0.494_{\pm 0.010}$ \\
                                                                                                                         & SGD & $70.74_{\pm 0.49}$ & $91.82_{\pm 0.10}$ & $1.230_{\pm 0.012}$ & $0.131_{\pm 0.004}$ & $0.427_{\pm 0.006}$ \\
                                                                                                                         \rowcolor{gray!10} \cellcolor{white}& IVON@mean & $72.67_{\pm 0.43}$ & $92.86_{\pm 0.14}$ & $1.118_{\pm 0.017}$ & $0.119_{\pm 0.002}$ & $0.397_{\pm 0.005}$ \\
                                                                                                                         \rowcolor{gray!10} \cellcolor{white}\multirow{-5}*{\begin{tabular}{l} \cellcolor{white} {DenseNet-121}\\ \cellcolor{white} (1M params) \end{tabular}} & IVON & ${\bf 73.68}_{\pm 0.37}$ & ${\bf 93.31}_{\pm 0.15}$ & ${\bf 0.940}_{\pm 0.012}$ & ${\bf 0.022}_{\pm 0.002}$ & ${\bf 0.361}_{\pm 0.004}$ \\
\cmidrule{1-7}
                                                                                                                                & AdamW & $65.88_{\pm 0.84}$ & $88.34_{\pm 0.56}$ & $2.893_{\pm 0.088}$ & $0.258_{\pm 0.006}$ & $0.578_{\pm 0.014}$ \\
                                                                                                                                & AdaHessian & $72.43_{\pm 0.36}$ & $91.92_{\pm 0.38}$ & $1.844_{\pm 0.044}$ & $0.194_{\pm 0.004}$ & $0.452_{\pm 0.008}$ \\
                                                                                                                                & SGD & $74.19_{\pm 0.11}$ & $92.41_{\pm 0.14}$ & $1.204_{\pm 0.012}$ & $0.137_{\pm 0.002}$ & $0.393_{\pm 0.004}$ \\
                                                                                                                                \rowcolor{gray!10} \cellcolor{white}& IVON@mean & $75.23_{\pm 0.23}$ & $93.45_{\pm 0.16}$ & $1.149_{\pm 0.010}$ & $0.136_{\pm 0.002}$ & $0.380_{\pm 0.003}$ \\
                                                                                                                                \rowcolor{gray!10} \cellcolor{white}\multirow{-5}*{\begin{tabular}{l} \cellcolor{white} {PreResNet-110}\\ \cellcolor{white} (deep, 4M params) \end{tabular}} & IVON & ${\bf 75.81}_{\pm 0.18}$ & ${\bf 93.93}_{\pm 0.19}$ & ${\bf 0.884}_{\pm 0.007}$ & ${\bf 0.030}_{\pm 0.003}$ & ${\bf 0.336}_{\pm 0.001}$ \\
\cmidrule{1-7}
                                                                                                                             & AdamW & $64.12_{\pm 0.43}$ & $86.85_{\pm 0.51}$ & $3.357_{\pm 0.071}$ & $0.278_{\pm 0.005}$ & $0.615_{\pm 0.008}$ \\
                                                                                                                             & AdaHessian & $56.42_{\pm 6.22}$ & $80.56_{\pm 4.81}$ & $2.503_{\pm 0.261}$ & $0.258_{\pm 0.014}$ & $0.666_{\pm 0.071}$ \\
                                                                                                                             & SGD & $74.46_{\pm 0.17}$ & $92.66_{\pm 0.06}$ & $1.083_{\pm 0.007}$ & $0.113_{\pm 0.001}$ & $0.376_{\pm 0.001}$ \\
                                                                                                                             \rowcolor{gray!10} \cellcolor{white}& IVON@mean & $74.51_{\pm 0.24}$ & $92.74_{\pm 0.19}$ & $1.284_{\pm 0.013}$ & $0.152_{\pm 0.003}$ & $0.399_{\pm 0.002}$ \\
                                                                                                                             \rowcolor{gray!10} \cellcolor{white}\multirow{-5}*{\begin{tabular}{l} \cellcolor{white} {ResNet-18}\\ \cellcolor{white} (wide, 11M params) \end{tabular}} & IVON & ${\bf 75.14}_{\pm 0.34}$ & ${\bf 93.30}_{\pm 0.19}$ & ${\bf 0.912}_{\pm 0.009}$ & ${\bf 0.021}_{\pm 0.003}$ & ${\bf 0.344}_{\pm 0.003}$ \\
\bottomrule
\end{tabular}
\caption{IVON results on CIFAR-100 compared with various baseline optimizers using convolutional networks with different widths and depths. IVON @mean denotes point estimate results evaluated at the mean of IVON posterior.}
\label{tab:resnets:cifar100}
\end{table}

\begin{table}[t!]\centering
\setlength{\tabcolsep}{3pt}

\begin{tabular}{llccccc}
\toprule
 &  & Acc. $\uparrow$ & Top-5 Acc. $\uparrow$ & NLL $\downarrow$ & ECE $\downarrow$ & Brier $\downarrow$ \\
\midrule
                                                                                                                        & AdamW & $46.62_{\pm 0.78}$ & $72.71_{\pm 0.75}$ & $2.387_{\pm 0.042}$ & $0.121_{\pm 0.004}$ & $0.692_{\pm 0.009}$ \\
                                                                                                                        & AdaHessian & $50.06_{\pm 0.53}$ & $76.09_{\pm 0.29}$ & $2.120_{\pm 0.016}$ & $0.084_{\pm 0.007}$ & $0.642_{\pm 0.004}$ \\
                                                                                                                        & SGD & ${\bf 51.08}_{\pm 0.22}$ & ${\bf 77.17}_{\pm 0.25}$ & ${\bf 1.989}_{\pm 0.007}$ & ${\bf 0.020}_{\pm 0.003}$ & ${\bf 0.622}_{\pm 0.002}$ \\
                                                                                                                        \rowcolor{gray!10} \cellcolor{white}& IVON@mean & ${\bf 50.71}_{\pm 0.38}$ & ${\bf 76.82}_{\pm 0.41}$ & $2.014_{\pm 0.017}$ & ${\bf 0.020}_{\pm 0.006}$ & ${\bf 0.629}_{\pm 0.005}$ \\
                                                                                                                        \rowcolor{gray!10} \cellcolor{white}\multirow{-5}*{\begin{tabular}{l} \cellcolor{white} {ResNet-20}\\ \cellcolor{white} (272k params) \end{tabular}} & IVON & ${\bf 50.85}_{\pm 0.42}$ & ${\bf 76.92}_{\pm 0.37}$ & $2.017_{\pm 0.016}$ & $0.060_{\pm 0.005}$ & $0.632_{\pm 0.004}$ \\
\cmidrule{1-7}
                                                                                                                         & AdamW & $50.01_{\pm 0.28}$ & $74.76_{\pm 0.32}$ & $5.515_{\pm 0.112}$ & $0.385_{\pm 0.003}$ & $0.851_{\pm 0.004}$ \\
                                                                                                                         & AdaHessian & $43.66_{\pm 10.76}$ & $69.86_{\pm 9.69}$ & $3.142_{\pm 0.320}$ & $0.189_{\pm 0.150}$ & $0.772_{\pm 0.044}$ \\
                                                                                                                         & SGD & $56.57_{\pm 1.00}$ & $80.46_{\pm 0.81}$ & $1.913_{\pm 0.056}$ & $0.126_{\pm 0.008}$ & $0.585_{\pm 0.012}$ \\
                                                                                                                         \rowcolor{gray!10} \cellcolor{white}& IVON@mean & ${\bf 58.47}_{\pm 0.10}$ & ${\bf 82.58}_{\pm 0.23}$ & $1.675_{\pm 0.008}$ & $0.046_{\pm 0.004}$ & ${\bf 0.542}_{\pm 0.003}$ \\
                                                                                                                         \rowcolor{gray!10} \cellcolor{white}\multirow{-5}*{\begin{tabular}{l} \cellcolor{white} {DenseNet-121}\\ \cellcolor{white} (1M params) \end{tabular}} & IVON & ${\bf 58.90}_{\pm 0.34}$ & ${\bf 82.69}_{\pm 0.35}$ & ${\bf 1.644}_{\pm 0.012}$ & ${\bf 0.035}_{\pm 0.002}$ & ${\bf 0.536}_{\pm 0.003}$ \\
\cmidrule{1-7}
                                                                                                                                & AdamW & $\smash{50.65_{\pm 0.0^{\ast}}}$ & $\smash{74.94_{\pm 0.0^{\ast}}}$ & $\smash{4.487_{\pm 0.0^{\ast}}}$ & $\smash{0.357_{\pm 0.0^{\ast}}}$ & $\smash{0.812_{\pm 0.0^{\ast}}}$ \\
                                                                                                                                & AdaHessian & $55.03_{\pm 0.53}$ & $78.49_{\pm 0.34}$ & $2.971_{\pm 0.064}$ & $0.272_{\pm 0.005}$ & $0.690_{\pm 0.008}$ \\
                                                                                                                                & SGD & $59.39_{\pm 0.50}$ & $81.34_{\pm 0.30}$ & $2.040_{\pm 0.040}$ & $0.176_{\pm 0.006}$ & $0.577_{\pm 0.007}$ \\
                                                                                                                                \rowcolor{gray!10} \cellcolor{white}& IVON@mean & ${\bf 60.85}_{\pm 0.39}$ & ${\bf 83.89}_{\pm 0.14}$ & $1.584_{\pm 0.009}$ & $0.053_{\pm 0.002}$ & ${\bf 0.514}_{\pm 0.003}$ \\
                                                                                                                                \rowcolor{gray!10} \cellcolor{white}\multirow{-5}*{\begin{tabular}{l} \cellcolor{white} {PreResNet-110}\\ \cellcolor{white} (deep, 4M params) \end{tabular}} & IVON & ${\bf 61.25}_{\pm 0.48}$ & ${\bf 84.13}_{\pm 0.17}$ & ${\bf 1.550}_{\pm 0.009}$ & ${\bf 0.049}_{\pm 0.002}$ & ${\bf 0.511}_{\pm 0.003}$ \\
\cmidrule{1-7}
                                                                                                                             & AdamW & $47.33_{\pm 0.90}$ & $71.54_{\pm 0.95}$ & $6.823_{\pm 0.235}$ & $0.421_{\pm 0.008}$ & $0.913_{\pm 0.018}$ \\
                                                                                                                             & AdaHessian & $51.80_{\pm 0.29}$ & $75.01_{\pm 0.10}$ & $3.416_{\pm 0.028}$ & $0.304_{\pm 0.002}$ & $0.748_{\pm 0.005}$ \\
                                                                                                                             & SGD & $61.39_{\pm 0.18}$ & $82.30_{\pm 0.22}$ & $1.811_{\pm 0.010}$ & $0.138_{\pm 0.002}$ & $0.536_{\pm 0.002}$ \\
                                                                                                                             \rowcolor{gray!10} \cellcolor{white}& IVON@mean & ${\bf 62.41}_{\pm 0.15}$ & ${\bf 83.77}_{\pm 0.18}$ & $1.776_{\pm 0.018}$ & $0.150_{\pm 0.005}$ & $0.532_{\pm 0.002}$ \\
\rowcolor{gray!10}\cellcolor{white}\multirow{-5}*{\begin{tabular}{l} \cellcolor{white} {ResNet-18}\\ \cellcolor{white} (wide, 11M params) \end{tabular}} & IVON & ${\bf 62.68}_{\pm 0.16}$ & ${\bf 84.12}_{\pm 0.24}$ & ${\bf 1.528}_{\pm 0.010}$ & ${\bf 0.019}_{\pm 0.004}$ & ${\bf 0.491}_{\pm 0.001}$ \\
\bottomrule
\end{tabular}
\caption{IVON results on TinyImageNet compared with various baseline optimizers using convolutional networks with different widths and depths. IVON @mean denotes point estimate results evaluated at the mean of IVON posterior. ($^*$) AdamW only converged for one of the five random seeds for PreResNet-110.}
\label{tab:resnets:tinyimagenet}
\end{table}

\subsection{Training with IVON for Image Classification}
\label{app:hyp2}
We train a ResNet-50 ($\approx 25.6$
million parameters)~\citep{HeZh16} with filter response normalization on the
ImageNet dataset ($\approx 1.2$ million examples with $1000$ classes)~\citep{DeDo09}.
Training for $200$ epochs takes around $30$ hours on $8$ A100 GPUs for all methods. Our distributed
implementation of IVON uses different random perturbations on each accelerator.
IVON's initial learning rate is $2.5$, we set $\beta_1=0.9$, $\beta_2 =
1 - 5 \cdot 10^{-6}$, $\delta = 5 \cdot 10^{-5}$, $h_0 = 0.05$ and
$\lambda = N = 1281167$. No clipping is used and we train with a
single MC sample. SGD uses a learning rate
of $0.5$ with same momentum $\beta_1=0.9$ and weight-decay $\delta = 5
\cdot 10^{-5}$. AdamW uses $\beta_1 = 0.9$, $\beta_2 = 0.999$,     
learning rate $0.001$ and weight-decay $0.1$. The damping parameter 
in AdamW is set to the PyTorch default value $\varepsilon=10^{-8}$, 
and larger values could potentially bring the AdamW performance closer to SGD.  
All methods anneal the learning rate to zero with a
cosine learning rate schedule after a linear warmup phase over 5
epochs.    

\revision{Here we also include additional image classification results using also deeper DenseNet-121 \citep{HuLi17} and ResNet-20 in addition to
ResNet-18 and PreResNet-110 \citep{HeZh16b} on CIFAR-10 and the previously reported CIFAR-100~\citep{krHi09} and TinyImageNet~\citep{LeYa2015}.}
The results are summarized in Tables~\ref{tab:resnets:cifar10},~\ref{tab:resnets:cifar100}~and~\ref{tab:resnets:tinyimagenet}. We also compare to AdaHessian~\citep{YaGh21}.
We find that IVON improves over other optimizers in terms of both accuracy and uncertainty\revision{, across all datasets and all metrics.}
Finally, IVON does not overfit on smaller datasets.   

\revision{For the experiments on CIFAR and TinyImageNet in~\Cref{tab:imagenet,tab:resnets:cifar10,tab:resnets:cifar100,tab:resnets:tinyimagenet},}
the hyperparameters of all methods were tuned only for the ResNet-20 on CIFAR-10, and kept fixed across the other models and datasets.
For SGD the learning rate $\alpha = 0.1$ was the largest stable learning rate across all models and datasets and gave the best results. AdaHessian uses $\alpha = 0.05$. 
It was not stable across all datasets when using the same learning rate as SGD as recommended by~\citet{YaGh21}.
AdamW uses learning rate $\alpha = 0.002$, except for the PreResNet-110 on TinyImageNet, where we reran with $\alpha = 0.0005$ to get it to converge.  We set $\beta_2 = 0.999$ in AdamW. 
IVON uses $\alpha = 0.2$, $\beta_2 = 1 - 10^{-5}$, $\lambda = N$ and $h_0 = 0.5$. All methods use gradient momentum $\beta_1 = 0.9$. We ran all optimizers
for $200$ epochs with batch-size $50$. The learning rate was warmed up for $5$ epochs using a linear schedule, and then decayed using a cosine
learning rate annealing.  The weight-decay is set to $\delta = 0.0002$
for all algorithms, datasets and models.

\subsection{In-domain and OOD Comparison to Bayesian Deep Learning Methods}
\label{app:hyp-bdl}
We train all ResNet-20 models with $200$ epochs and batch size $50$. 
Weight decay is set to $0.0002$.
Apart from SWAG, which requires custom scheduling, 
all other methods use $5$ warm-up epochs followed by a cosine annealing learning rate schedule that decays to zero. 
We do $5$ runs with different random seeds and report the average results and their standard deviations in the tables. 

For the uncertainty estimation metrics used in in-domain and distributional shift experiments, 
we follow \citet{SnOv19} and report \revision{three} metrics: negative log-likelihood (NLL), expected calibration error (ECE), and Brier score. 
For the OOD experiments we used the same metrics as \citet{LiLi18}, i.e.\ False Positive Rate (FPR), the share of misclassified OOD samples, at 95\% TPR, detection error, which 
measures the probability of misclassifications for 95\% TPR,
Area Under the Receiver Operating Characteristic curve (AUROC), AUPR-in, and AUPR-out.
Here, AUPR stands for Area under the Precision-Recall for the in-domain data (AUPR-in) or OOD data (AUPR-out), respectively.

The specific training hyperparameters for each method are: 
\begin{itemize}
	\item SGD and IVON use the same setting as in Section~\ref{app:hyp2}, except that SGD also uses learning rate $0.2$ which is stable for ResNet-20;
	\item VOGN uses the same hyperparameter setup as IVON;
	\item BBB uses learning rate $0.002$. We set the same initial posterior as IVON and train BBB without using a cold posterior;
	\item Laplace uses the linearized last-layer Laplace implementation from \citet{DaKr21} and the prior precision is set to $10.0$ corresponding to the same prior setup as other methods;
	\item MC dropout uses learning rate $0.2$ and a fixed \revision{dropout} rate of $0.05$;
    \item For SWAG, we first do normal training with cosine annealing from lr $0.05$ to $0.01$ over $160$ epochs, then do $40$ SWAG epochs with constant \revision{learning rate}
     $0.01$ and maintain a rank $20$ approximation 
    of the SWAG posterior as is done in~\citep{MaIz19}.
\end{itemize}

We use $64$ posterior samples for IVON, BBB, Laplace, SWAG and VOGN. 
For MC dropout, we only draw $32$ samples for all experiments as we observe no improvement when drawing $64$ samples.

\subsection{Finetuning on GLUE}
\label{app:roberta}
\begin{table}[ht]\centering

    \setlength{\tabcolsep}{3pt}
    
    \begin{tabular}{lcccccccc}
    \toprule
     &  MNLI-m & QNLI & QQP & RTE & SST2 & MRPC & CoLA & STS-B \\
     Metric & Acc. & Acc. & Acc. & Acc. & Acc. & Acc / F1 & Spearman & MCC \\
     \#Train & 393k & 105k & 364k & 2.5k & 67k & 3.7k & 8.5k & 7k \\
     \#Validation & 9.8k & 5.5k & 40.4k & 277 & 872 & 408 & 1k & 1.5k \\
     \bottomrule
    \end{tabular}
    \caption{Dataset sizes of individual GLUE tasks used in this paper and the used evaluation metrics.}
\end{table}
GLUE~\citep{WaSi18} is a multi-task benchmark consisting of in total $9$ diverse tasks which capture classification and regression problems.
We use all tasks but WNLI~\citep{LeDa12} following previous work~\citep{DeCha18}.
Namely, we use: CoLA~\citep{WaSi18_2}, MNLI~\citep{WiNa18}, MRPC~\citep{DoBro05}, QNLI~\citep{WaSi18}, QQP, RTE, SST2~\citep{SoPe13}, and STS-B~\citep{CeDi17}.

For IVON, we use the same hyperparameters for the two models used in our experiments: RoBERTa and DeBERTav3 shown in~\Cref{subsec:benefit-finetune-mlm}.
We use an initial learning rate of $0.1$ or $0.2$ which is decayed to $0.0$ using cosine decay.
We set $\beta_1=0.9$, $\beta_2 = 1-10^{-5}$, $h_0 = 1.0$, a weight decay factor of $10^{-5}$, and also use element-wise clipping of $10^{-3}$.
Furthermore, we use $500$ warmup steps.

For RoBERTa \revision{with AdamW}, we use the hyperparameters reported in~\citep[Table 10]{liuOtt19}.
Namely, we sweep learning rates over $\{10^{-5}, 2\cdot 10^{-5}, 3 \cdot 10^{-5}\}$.
We use a weight decay of $0.1$, $\beta_1=0.9$, and $\beta_2=0.98$.

For DeBERTAv3 \revision{with AdamW} we use the hyperparameters as reported in~\citep[Table 11]{HeGa23} but were unable to sweep all possible combinations that are listed due to the high computational demand.
Therefore, we fix the number of warmup steps to $500$ and the batch size to $32$.
Also, we do not use last layer dropout.
We sweep learning rates over $\{5\cdot 10^{-6}, 8\cdot 10^{-6}, 9\cdot 10^{-6}, 10^{-5}\}$, use a weight decay of $0.1$, $\beta_1=0.9$, and $\beta_2=0.999$.

We evaluate after each epoch and train for up to $10$ epochs on every dataset but MRPC, where we allow $15$ epochs.
The batch size is always set to $32$ for both AdamW and IVON.

\subsection{Predicting Generalization and Understanding Models' Sensitivity to Data}
\label{app:mpe}
\label{app:sensitivity-method}
We {use} the memory-perturbation equation (MPE) by
\citet{NiXu23}.
In their framework, prediction error and variances for a multi-output vector $\vf_i(\vparam_t) \in \real^C$ (with $C$ being the number of classes) at iteration $t$ are obtained as follows,
\[ 
   \ve_{it} = \softmax(\vf_i(\text{\vtheta}_t)) - \vy_i, \qquad
   \vV_{it} =\nabla\vf_i(\vtheta_t)^\top \text{diag}(\vsigma^2_{t}) \nabla\vf_i(\vtheta_t),
\] 
where $\softmax(\cdot)$ is the softmax function and $\nabla\vf_i(\vtheta_t) \in \mathbb{R}^{P \times C}$ is
the Jacobian with $P$ being the number of parameters.

For IVON, we set $\vparam_t = \vm_t$ and use the posterior variance $\vsigma^2_t = 1 / \lambda (\vh_t + \delta)$. For SGD and AdamW, we construct $\vsigma^2_{t}$ in ad-hoc ways. For SGD we use $\vsigma^2_{t} = 1/N(\bm{1} + \delta)$. For AdamW we use $\vsigma^2_{t} = 1/N(\sqrt{\vh_t} + \delta )$, where $\vh_t$ is the second moment vector that maintains a running-average of squared gradients.

For all data sensitivity experiments in the main paper we used the following hyperparameters to train a ResNet-50 on ImageNet for $100$ epochs. 
IVON uses an initial learning rate of $3$, $\beta_1=0.9$, $\beta_2 =
1-10^{-6}$, $h_0 = 0.008$, $\lambda = N$ and a weight decay of $\delta
= 5 \cdot 10^{-5}$. $h_0$ was selected on a grid of $[0.008, 0.01, 0.05]$ to achieve a faithful estimate of generalization performance while keeping a competitive test accuracy. 
AdamW uses a learning rate of $0.001$, $\beta_1=0.9$, $\beta_2=0.999$
and weight decay $0.1$. 
Both methods use $5$ warmup epochs after which the learning rate is
decayed to $0$ using cosine decay. The model trained with IVON has an
accuracy of $75\%$, whereas the AdamW model has $74.7\%$ accuracy.

\section{Additional Results}
\label{app:additional-results}

\begin{table}[t!]\centering
\setlength{\tabcolsep}{3pt}

\begin{tabular}{llccccc}
\toprule
 &  & Acc. $\uparrow$	  & NLL $\downarrow$ 	   & ECE $\downarrow$ 	& Brier $\downarrow$ & AUROC $\uparrow$ \\
\midrule

& AdamW & $64.35_{\pm 0.27}$ & $0.666_{\pm 0.026}$ & $0.322_{\pm 0.026}$ & $0.658_{\pm 0.048}$ &${\bf 0.579}_{\pm 0.019}$ \\
& SGD 	& ${\bf 67.65}_{\pm 0.92}$ & $ 0.631_{\pm 0.011}$ & $0.060_{\pm 0.006}$ & $0.448_{\pm 0.006}$ & $0.548_{\pm 0.027}$ \\
\rowcolor{gray!10} \cellcolor{white}& IVON@mean &  ${\bf 68.54}_{\pm 0.00}$ & ${\bf 0.623}_{\pm 0.03}$ & ${\bf 0.030}_{\pm 0.005}$ & ${\bf 0.432}_{\pm 0.003}$ & $0.509_{\pm 0.041}$ \\
\rowcolor{gray!10} \cellcolor{white}\multirow{-4}*{\begin{tabular}{l} \cellcolor{white} {CoLA} \end{tabular}}
& IVON & ${\bf 68.54}_{\pm 0.00}$ & ${\bf 0.623}_{\pm 0.03}$ & ${\bf 0.029}_{\pm 0.005}$ & ${\bf 0.432}_{\pm 0.003}$ & $0.510_{\pm 0.041}$ \\

\cmidrule{1-7}

& AdamW & $84.66_{\pm 0.46}$ & $1.929_{\pm 0.344}$ & $0.136_{\pm 0.007}$ & $0.285_{\pm 0.011}$ & $0.725_{\pm 0.013}$ \\
& SGD & $85.06_{\pm 0.38}$ & ${\bf 0.468}_{\pm 0.013}$ & ${\bf 0.065}_{\pm 0.009}$ & $0.233_{\pm 0.011}$ & $0.764_{\pm 0.016}$ \\
\rowcolor{gray!10} \cellcolor{white}& IVON@mean & ${\bf 89.55}_{\pm 0.20}$ & ${\bf 0.428}_{\pm 0.023}$ & ${\bf 0.061}_{\pm 0.003}$ & $0.251_{\pm 0.023}$ & $ {\bf0.811}_{\pm 0.025}$ \\
\rowcolor{gray!10} \cellcolor{white}\multirow{-4}*{\begin{tabular}{l} \cellcolor{white} {IMDB} \cellcolor{white} \end{tabular}} 
& IVON & $ 87.73_{\pm 0.96}$ & $0.568_{\pm 0.119}$ & ${\bf 0.065}_{\pm 0.010}$ & ${\bf 0.199}_{\pm 0.018}$ & $0.751_{\pm 0.031}$ \\

\cmidrule{1-7}

& AdamW & $90.65_{\pm 0.32}$ & $0.985_{\pm 0.046}$ & $0.408_{\pm 0.031}$ & $ 0.171_{\pm 0.006}$ & $0.826_{\pm 0.006}$ \\
& SGD & $89.57_{\pm 0.40}$ & $0.386_{\pm 0.009}$ & $0.055_{\pm 0.004}$ & $0.167_{\pm 0.005}$ & $0.845_{\pm 0.004}$ \\
\rowcolor{gray!10} \cellcolor{white}& IVON@mean &  ${\bf 92.43}_{\pm 0.01}$ & ${\bf 0.233}_{\pm 0.003}$ & ${\bf 0.017}_{\pm 0.002}$ & ${\bf 0.118}_{\pm 0.001}$ & ${\bf 0.871}_{\pm 0.003}$ \\
\rowcolor{gray!10} \cellcolor{white}\multirow{-4}*{\begin{tabular}{l} \cellcolor{white} {AG News} \cellcolor{white}  \end{tabular}} & IVON & ${\bf 92.46}_{\pm 0.01}$ & ${\bf 0.231}_{\pm 0.002}$ & ${\bf 0.014}_{\pm 0.003}$ & ${\bf 0.117}_{\pm 0.001}$ & ${\bf 0.871}_{\pm 0.002}$ \\

\bottomrule
\end{tabular}
\caption{IVON results on NLP classification datasets compared with to SGD and AdamW. IVON@mean denotes point estimate results evaluated at the mean of IVON posterior.}
\label{tab:rnn_classification}
\end{table}

\subsection{IVON with Recurrent Neural Networks}
\label{app:rnn}
We train a simple
model based on Gated Recurrent Units~\citep{ChVa14} on three text
classification datasets (CoLA, IMDB and AG News).
The model consists of an embedding layer, two GRUs and a fully connected layer, for a total of 2 million parameters. 
We train the same model with SGD, AdamW and IVON. IVON results are
evaluated both at the mean and at a Monte-Carlo approximation of the
posterior using 64 samples. Results are reported in
~\Cref{tab:rnn_classification}. IVON improves both accuracy
and uncertainty compared to the baselines. The chosen model can easily
overfit the presented datasets, achieving close to $100\%$ accuracy on
the training set. Therefore, extra care is required when choosing the
hyperparameters for AdamW and, especially, SGD. However, we find it easier to tune IVON for satisfactory results both in terms of accuracy and uncertainty.

\subsection{Robustness to Distribution Shift} 
\label{app:distshift}

\begin{figure}[t!]
	\centering
	\begin{tabular}{cc}
		\includegraphics[width=0.47\linewidth]{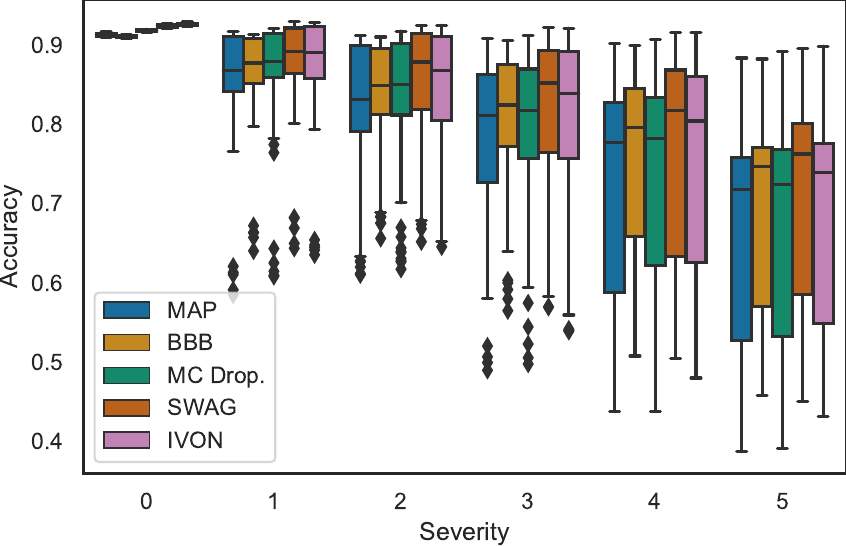} &
		\includegraphics[width=0.47\linewidth]{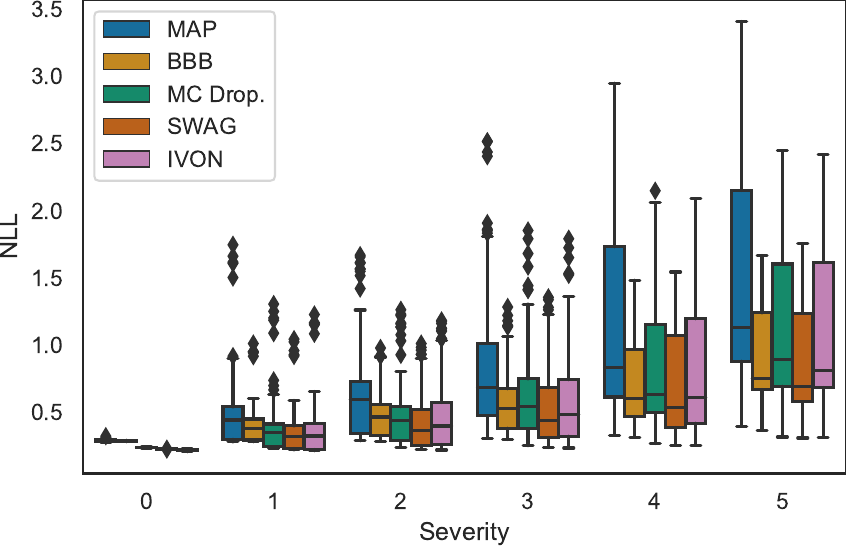} \\
		\includegraphics[width=0.47\linewidth]{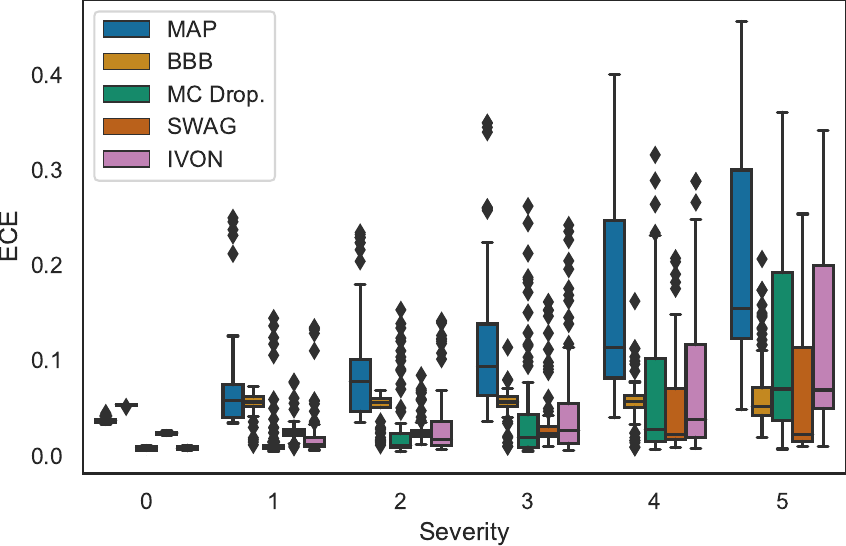} & \includegraphics[width=0.47\linewidth]{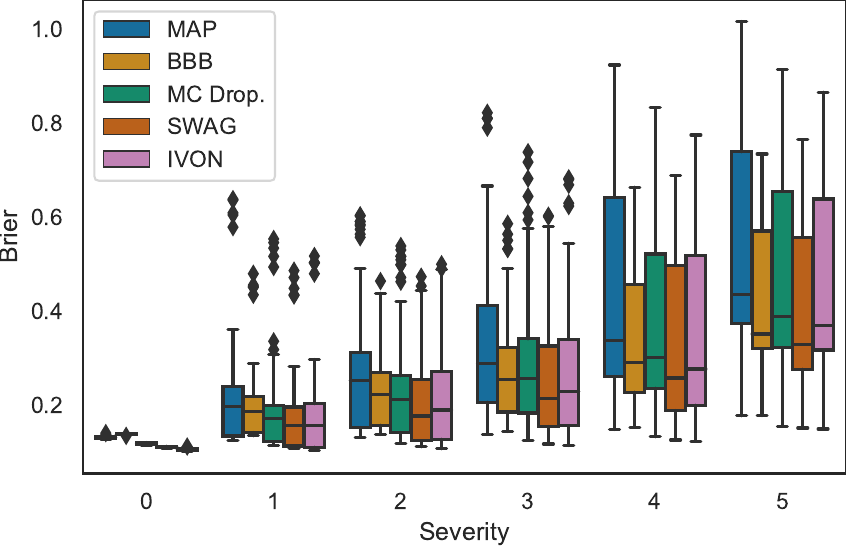} \\
	\end{tabular}
	\caption{Distributional shift results on CIFAR-10-C with various degree of severities. Severity 0 corresponds to the in-domain case.}
	\label{fig:distshift}
      \end{figure}

Having trained and evaluated various models on CIFAR-10 in the
in-domain scenario, here we conduct distributional shift experiments,
where we use the previously trained networks to directly classify
CIFAR-10 test set images corrupted with artificial perturbations. For
this we use the CIFAR-10-C \citep{HeDi19} dataset which collects a
range of common image distortions and artifacts, each with 5 severity
levels. The results are grouped by severity level and summarized in~\cref{fig:distshift} on the next page.

\begin{figure}[b!]
\centering
\begin{minipage}[t]{0.30\linewidth}
    \centering
    \raisebox{0.2in}{\includegraphics[width=\linewidth]{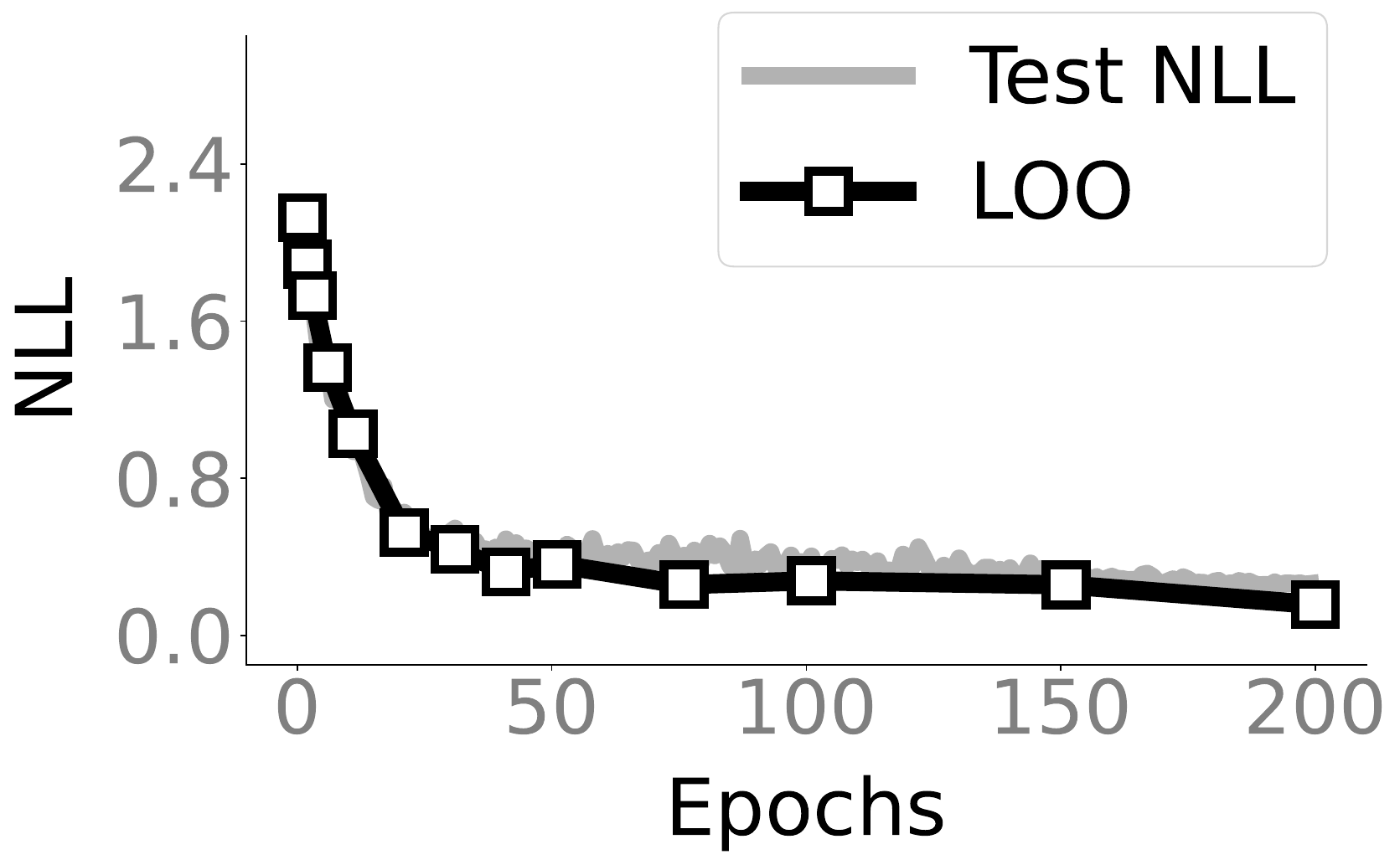}}
    \vspace*{-0.465in}
\end{minipage}
\hfill
\begin{minipage}[t]{0.28\linewidth}
    \centering
    \raisebox{0.2in}{\includegraphics[width=\linewidth]{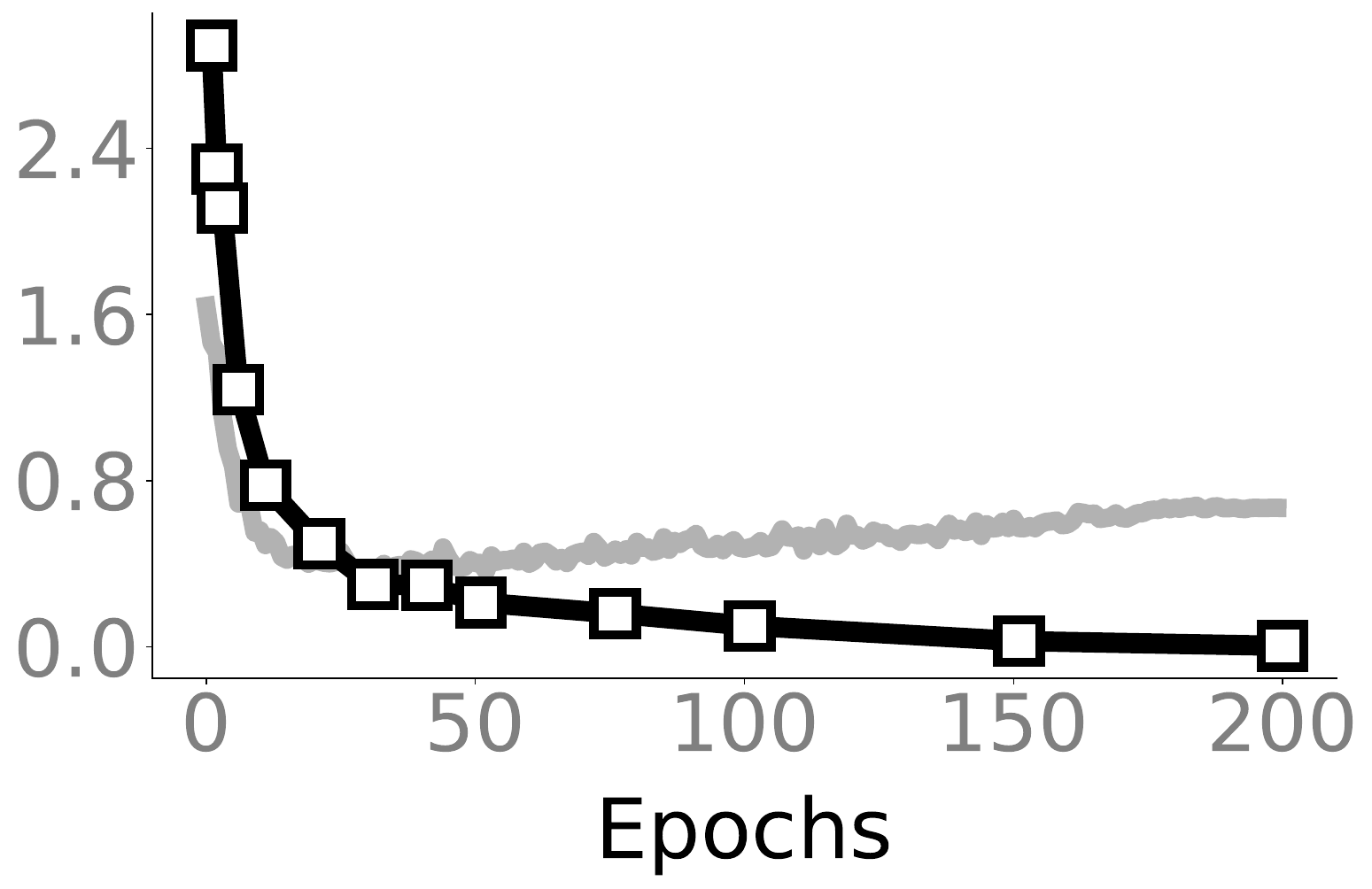}}
    \vspace*{-0.465in}
\end{minipage}
\hfill
\begin{minipage}[t]{0.28\linewidth}
    \centering
    \raisebox{0.2in}{\includegraphics[width=\linewidth]{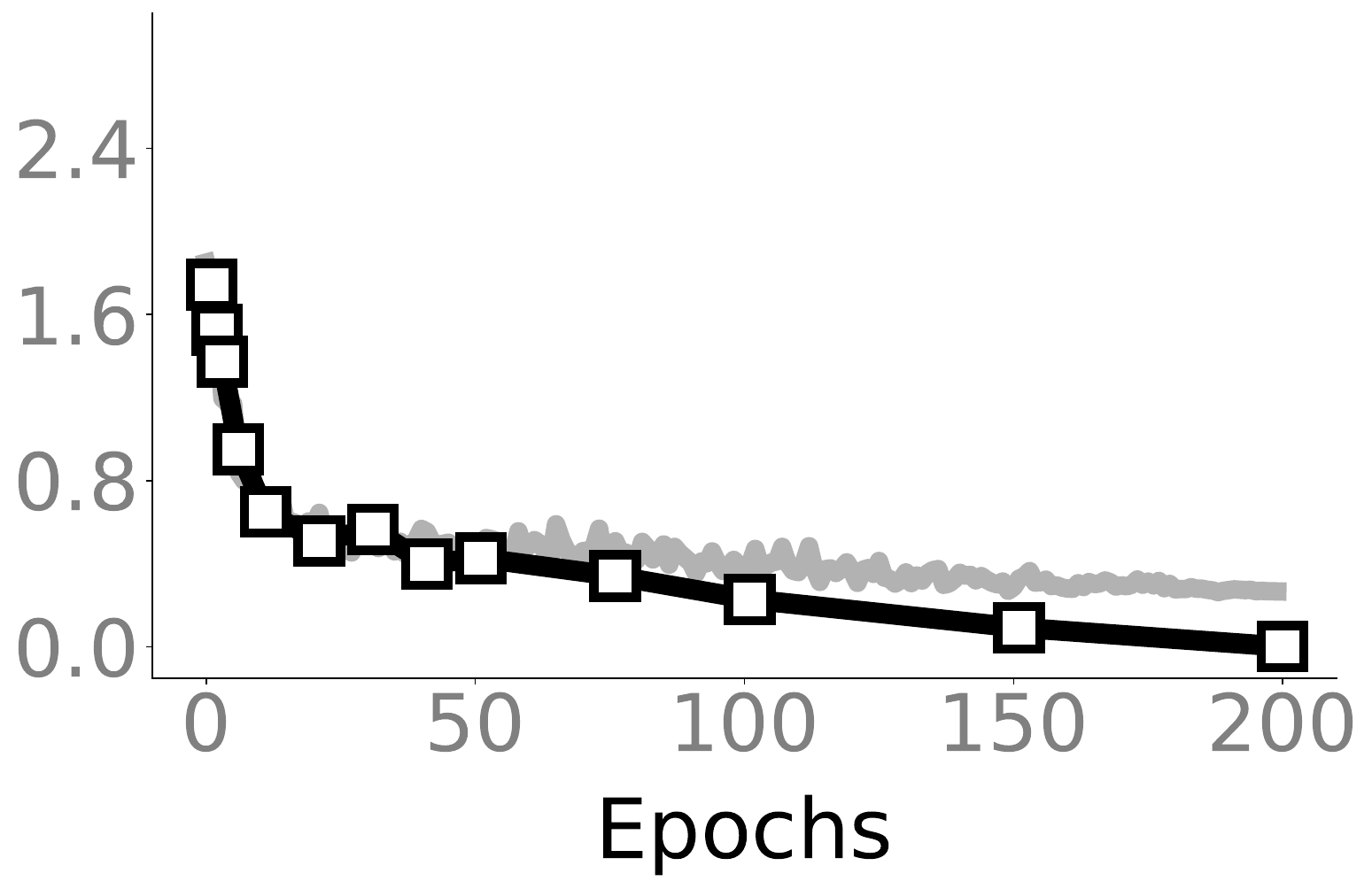}}
    \vspace*{-0.465in}
\end{minipage}
\centering
\begin{minipage}[t]{0.30\linewidth}
    \centering
    \raisebox{0.2in}{\includegraphics[width=\linewidth]{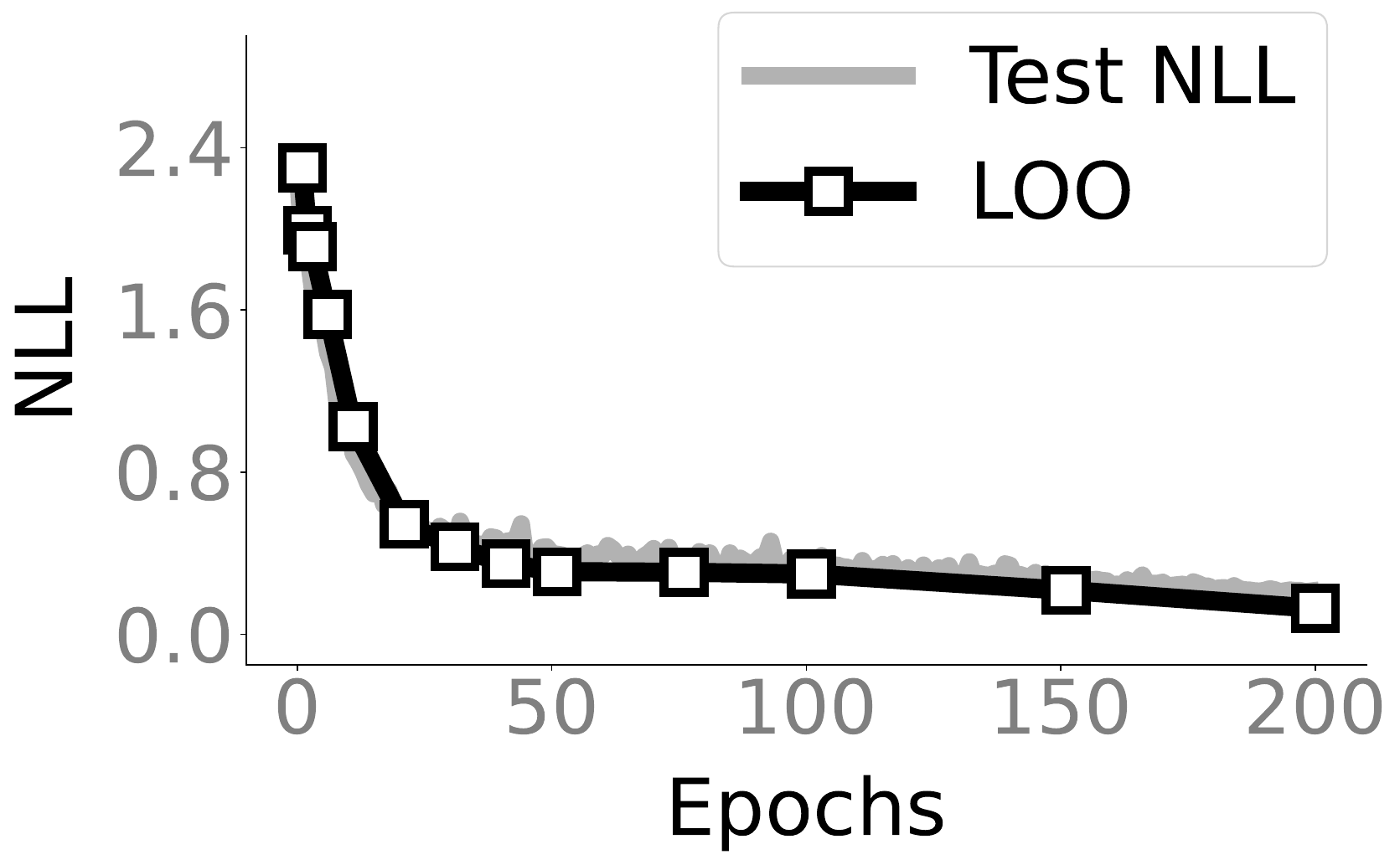}}
    \vspace*{-0.465in}
\end{minipage}
\hfill
\begin{minipage}[t]{0.28\linewidth}
    \centering
    \raisebox{0.2in}{\includegraphics[width=\linewidth]{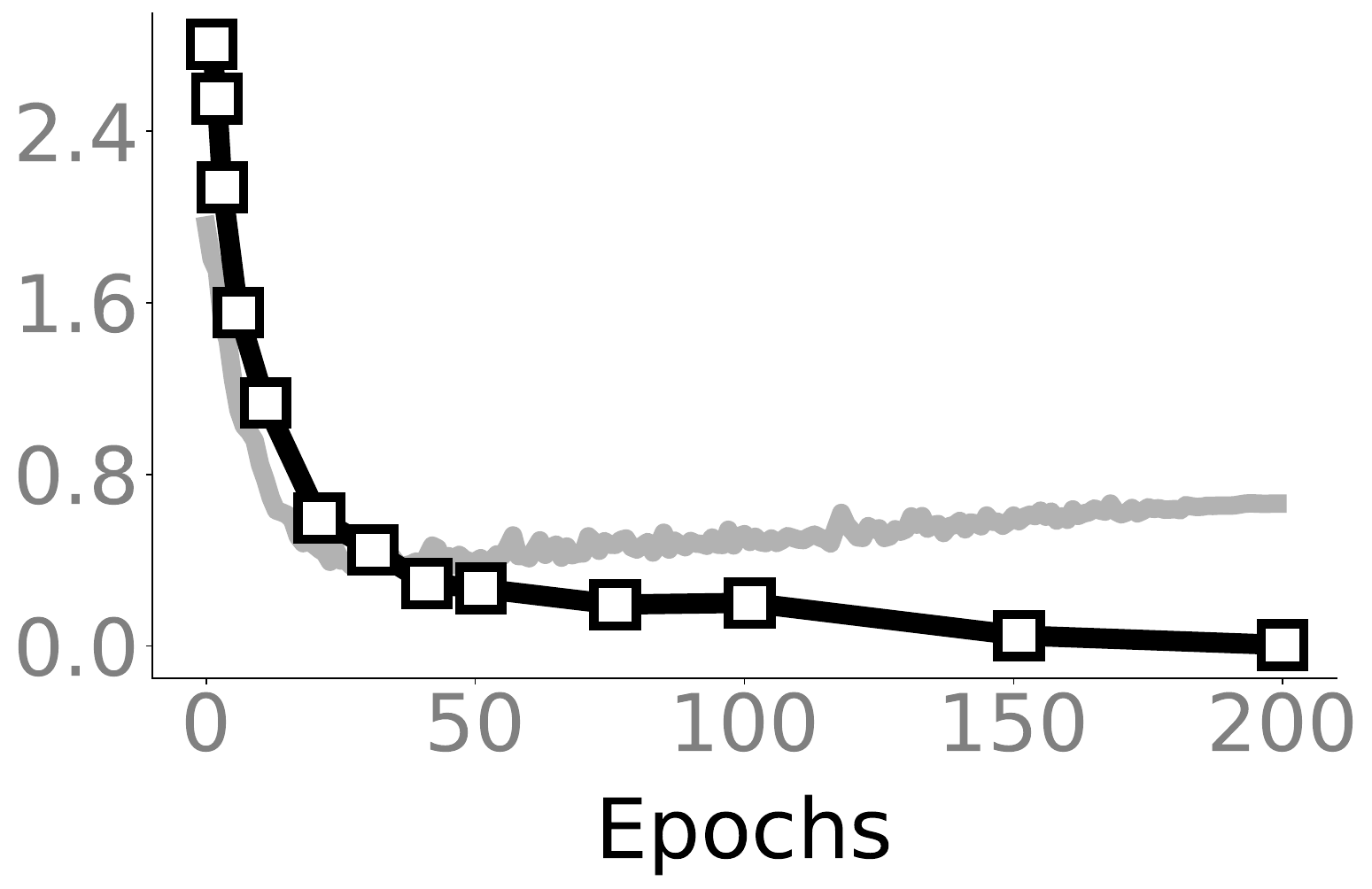}}
    \vspace*{-0.465in}
\end{minipage}
\hfill
\begin{minipage}[t]{0.28\linewidth}
    \centering
    \raisebox{0.2in}{\includegraphics[width=\linewidth]{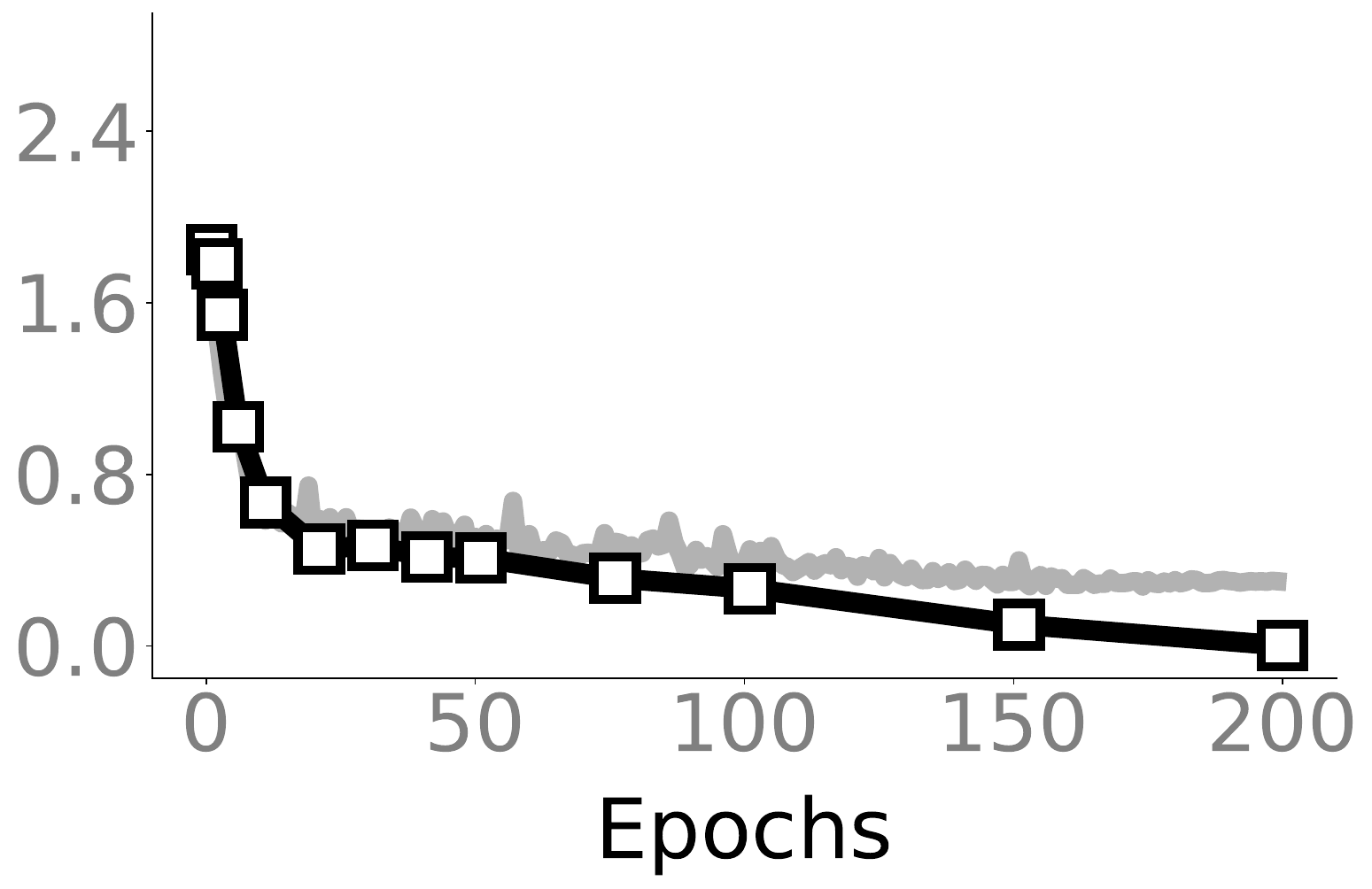}}
    \vspace*{-0.465in}
\end{minipage}
\caption{We predict test NLL using LOO estimation during training of two models on CIFAR10. From top to bottom the models are:
ResNet--18 and PreResNet--110. IVON (first column) allows us to faithfully predict generalization, while the heuristic LOO estimates with AdamW (second column) and SGD (third column) work less well. \label{fig:sensitivity_cifar10}}
\end{figure}

In general, the performance of all models decrease with increasing severity, as the classification task is getting harder. We observe that IVON keeps the best performance for low severity levels. For high severity levels, IVON is notably outperformed by SWAG. Despite this, IVON in general is still comparable to BDL baselines for high severity cases. And as an optimizer, it remains a better choice over the standard SGD training.

\revision{
  \subsection{NeurIPS 2021 Approximate Inference Competition}
  \label{app:challenge}
An earlier version of IVON won the first place\footnote{For the
  official results, see 
  \href{https://izmailovpavel.github.io/neurips_bdl_competition}{https://izmailovpavel.github.io/neurips\_bdl\_competition}.}
in both the light and extended track of the NeurIPS 2021
Competition on Approximate Inference in Bayesian Deep
Learning~\citep{WiIz22}. The earlier version included an additional heuristic damping to $\widehat \vh$ in~line~3 of~\Cref{alg:ivon}
and the weight-decay was added in line~4 rather than in lines~5~and~6. We found the damping term to be unnecessary
when using a proper Hessian initialization $h_0$ and momentum $\beta_2$ and therefore removed it, making IVON easier to tune.
The three highest scoring submissions to the competition are summarized in~\Cref{tab:bdlchallenge}. 
First place is Multi-IVON (using the earlier version of IVON), which is a mixture-of-Gaussian ensemble (with uniform weights) as described
in the experiments section on uncertainty estimation in the main paper. The second place solution (Multi-SWAG) uses
multiple runs of SWAG to construct a mixture-of-Gaussian
approximation~\citep{IzVi21} with SGLD~\citep{WeTe11} as a base
optimizer. Third place was obtained by a deep ensembling method called
sequential anchored ensembles (SAE)~\citep{DeLo21}.
In~\Cref{tab:bdlchallenge}, 'Agree' denotes predictive agreement with a ground-truth Bayesian posterior obtained by running Hamiltonian Monte-Carlo method on hundreds of TPUs.
TVD denotes the total variation distance and W$_2$ the Wasserstein-2 distance between this ground-truth predictive posterior and the approximate posterior. We refer to~\citet{WiIz22} for more details.
}

\begin{table}[h!]
  \centering
  \begin{tabular}{llcccccc}
    \toprule
    \multirow{2}{1cm}{Rank} & \multirow{2}{1cm}{Method} & \multicolumn{2}{c}{CIFAR-10} & \multicolumn{2}{c}{MedMNIST} & UCI \\
     & & Agree $\uparrow$ & TVD $\downarrow$ & Agree $\uparrow$ & TVD $\downarrow$ & W$_2$ $\downarrow$ \\
    \midrule 
    1 & Multi-IVON$^\dagger$ & \textbf{78.7\%} & \textbf{0.198} & 88.4\% & 0.099 & \textbf{0.094} \\
    2 & Multi-SWAG & 77.8\% & 0.219 & \textbf{89.0\%} & \textbf{0.098} & 0.166 \\
    3 & SAE & 77.3\% & 0.210 & 87.5\% & 0.107 & 0.116 \\ 
     \midrule
     \rowcolor{gray!10}   & Multi-IVON (\Cref{alg:ivon}) & 78.2\% & 0.204 & 89.1\% & 0.097 & 0.075 \\
      \bottomrule
  \end{tabular}
\caption{An earlier version of IVON (denoted by $\dagger$) won the NeurIPS 2021 competition
  on approximate inference in Bayesian deep learning~\citep{WiIz22}. The second best method used a combination of SWAG and SGLD. 
  Third place was a Sequential Anchored Ensemble (SAE). In the last row of the table we also report results achieved with~\Cref{alg:ivon} which performs 
  similarly well as the previous version of IVON. \label{tab:bdlchallenge}}
\end{table}

\subsection{Predicting Generalization}
\label{app:exp-predict-gen}
In~\Cref{fig:sensitivity_cifar10}, we conduct additional experiments with ResNet-18 and PreResNet-10 on the CIFAR-10 dataset. We estimate generalization performance during training using the LOO criterion described in~\Cref{app:sensitivity-method}. The accuracy of IVON is similar to the SGD baseline. IVON however results in a more faithful estimate of the generalization performance in comparison to AdamW and SGD. We evaluate the sensitivity of the models to data perturbation as described in~\Cref{app:sensitivity-method} with the difference that we compute the sensitivities as $\ve_i\vv_i$, where $\vv_i$
are the diagonal elements of $\vV_i$ and $\vv_i \ve_i$ is an element-wise product. For training the models, we use the same hyperparameters as in the image classification experiments. 
The exception is the Hessian initialization $h_0$. We do a grid search over the values $[0.01, 0.05, 0.1, 0.5]$. We select the value that results in a faithful estimate of the generalization performance while keeping a competitive test accuracy. The Hessian initialization is set to $h_0=0.1$ for both models. The test accuracies are $93.68\%$ for ResNet--18 and $93.52\%$ for PreResNet--110 with predictions at the mean of the variational posterior.

\end{document}